\documentclass[10pt,letterpaper]{article}
\usepackage[top=0.85in,left=2.75in,footskip=0.75in]{geometry}

\usepackage{hyperref}
\usepackage{amsmath,amssymb}
\usepackage{algorithmic}
\usepackage{algorithm}

\usepackage{changepage}
\usepackage{xfp}
\usepackage{bm}
\usepackage[final]{changes}
\definechangesauthor[name={hdh}, color=orange]{per}
\usepackage[utf8x]{inputenc}
\usepackage{subcaption}
\usepackage{ulem}
\usepackage{textcomp,marvosym}

\usepackage{cite}

\usepackage{nameref,hyperref}
\usepackage[T1]{fontenc}
\usepackage[right]{lineno}
\usepackage{colortbl}
\usepackage{microtype}
\DisableLigatures[f]{encoding = *, family = *}

\usepackage{array}

\newcolumntype{+}{!{\vrule width 2pt}}

\newlength\savedwidth

\raggedright
\usepackage[aboveskip=1pt,labelfont=bf,labelsep=period,justification=raggedright,singlelinecheck=off]{caption}

\definecolor{tgcolor}{rgb}{0.8,0.2,0.2}

\makeatletter
\renewcommand{\@biblabel}[1]{\quad#1.}
\makeatother

\usepackage{lastpage,fancyhdr,graphicx}
\usepackage{epstopdf}
\fancyheadoffset[L]{2.25in}
\fancyfootoffset[L]{2.25in}
\begin{document}
\vspace*{0.2in}

\begin{flushleft}
{\Large
\textbf\newline{Latent representation prediction networks}
}
\newline
\\
Hlynur Davíð Hlynsson\textsuperscript{*},
Merlin Schüler\textsuperscript{},
Robin Schiewer\textsuperscript{},
Tobias Glasmachers\textsuperscript{},
Laurenz Wiskott\textsuperscript{}
\bigskip
\\
Institut für Neuroinformatik, Ruhr-Universität Bochum, Bochum, Germany
\\
\bigskip

* \href{mailto:hlynur.hlynsson@ini.rub.de}{hlynur.hlynsson@ini.rub.de}

\end{flushleft}
\section*{Abstract}

Modern reinforcement learning methods for planning often incorporate an unsupervised learning step where the training objective leverages only static inputs, such as reconstructing observations. These representations are then combined with predictor functions for simulating rollouts to navigate the environment.  We are interested in advancing this idea by taking advantage of the fact that we are navigating a dynamic environment with visual stimulus and would like a representation that is specifically designed for control and action in mind. We propose to simultaneously learn a feature map that is maximally predictable for a predictor function. This results in representations that are well-suited, by design, for the downstream task of planning, where the predictor is used as an approximate forward model.

To this end, we introduce a new way of jointly learning this representation along with the prediction function, a system we dub Latent Representation Prediction Network (LARP). The prediction function is used as a forward model for a search on a graph in a viewpoint-matching task and the representation learned to maximize predictability is found to outperform a pre-trained representation. The sample-efficiency \added{and overall performance} of our approach is shown to rival standard reinforcement learning methods and our learned representation transfers successfully to novel environments.

\section*{Introduction}

While modern reinforcement learning algorithms reach super-human performance on tasks such as gameplaying, they remain woefully sample inefficient compared to humans. An algorithm that is data~\cite{icpram19} or sample-efficient \cite{wang2016sample} requires only  few samples for good performance and the study of sample-efficient control is currently an active research area \cite{corneil2018efficient}~\cite{du2019good}~\cite{saphal2020seerl}~\cite{NIPS2018_8044}. A powerful tool for increasing the sample-efficiency of machine learning methods is dimensionality reduction. 

 There has been much recent work on methods that take advantage of compact, low-dimensional representations of states for search and exploration~\cite{kurutach2018learning}~\cite{corneil2018efficient}~\cite{{xu1029regression}}. One of the advantages of this approach is that a good representation aids in faster and more accurate planning.
This holds in particular when the latent space is of much lower dimensionality than the state space\cite{hamilton2014efficient}. For high-dimensional inputs, such as image data, a representation function is frequently learned to reduce the complexity for a controller.

\begin{figure}[htb]
\begin{adjustwidth}{-0.in}{0in}
\caption{{\bf Conceptual overview of our method.} The important components are the representation network along with the predictor network. Together, they comprise a LARP network which is utilized by a planning algorithm.}
\centering
\includegraphics[width=0.7\textwidth]{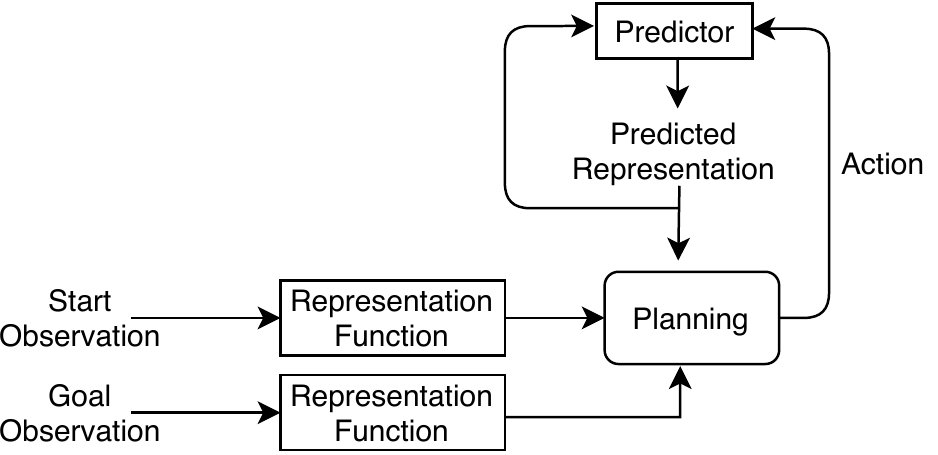}
\label{Conceptual1.pdf}
\end{adjustwidth}

\end{figure}

In deep reinforcement learning, the representation and the controller are learned simultaneously. Similarly, a representation can in principle be learned along with a forward model for classical planning in high-dimensional space. This paper introduces \textit{Latent Representation Prediction} (LARP) networks, a novel neural network-based method for learning a domain representation and a transition function for planning within the learned latent space (Fig.~\ref{Conceptual1.pdf}). During training, the representation and the predictor are learned simultaneously from transitions in a \added{self-}supervised manner. We train the predictor to predict the most likely future representation, given a current representation and an action. The predictor is then used for planning by navigating the latent space defined by the representation to reach a goal state.

Optimizing control in this manner after learning an environment model has the advantage of allowing for learning new reward functions in a fast and sample-efficient manner.
\added{After the representation is learned, we find said goal state by conventional path planning.} Disentangling the reward from the transition function in such a way is helpful when learning for multiple or changing reward functions and aids with learning when there is no reward available at all. Thus, it is also good for a sparse or a delayed-reward setting.

A problem that can arise in representation learning is the one of trivial features. This can happen when the method is optimizing an objective function that has a straightforward, but useless, solution. For example, Slow Feature Analysis (SFA) \cite{wiskott2002slow} has the objective of extracting the features of time series data that vary the least with time. This is easily fulfilled by constant functions, so SFA includes an optimization constraint to avoid constant solutions.

Constant features would in a similar manner be maximally predictable representations for our system. Therefore, we study three different approaches to prevent this trivial representation from being learned, we either: \textbf{(i)} design the architecture such that the output is sphered, \textbf{(ii)} regularize it with a contrastive loss term, or \textbf{(iii)} include a reconstruction loss term along with an additional decoder module. 


We compare these approaches and validate our method experimentally on a visual environment: a  viewpoint-matching task using the NORB data set \cite{lecun2004learning}, where the agent is presented with a starting viewpoint of an object and the task is to produce a sequence of actions such that the agent ends up with the goal viewpoint. As the NORB data set is embeddable on a cylinder \cite{schuler2018gradient} \cite{hadsell2006dimensionality} or a sphere \cite{wang2018toybox}, we can visualize the actions as traversing the embedded manifold. \added{ Our approach compares favorably to state-of-the-art methods on our testbed with respect to sample efficiency, but our asymptotic performance is still outclassed by other approaches.}

\section*{Related work}

Most of the related work falls into the categories of reinforcement learning, visual planning, or representation learning.  The primary \replaced{difference between ours and}{conceptual shortcoming of}  other model-based methods is that the representation is learned by optimizing auxiliary objectives which are \replaced{not directly useful for solving the main task.}{of direct interest for maximizing rewards.}  

\subsection*{Reinforcement learning}
There are many works in the literature that also approximate the transition function of environments, for instance by performing explicit latent-space planning computations \cite{tamar2016value}
\cite{srinivas2018universal} \cite{hafner2018learning} \cite{henaff2017model} \cite{chua2018deep} \cite{gal2016improving} as part of learning and executing policies. Gelada et al.\ \cite{gelada2019deepmdp} train a reinforcement learning (RL) agent to simultaneously predict rewards as well as future latent states. Our work is distinct from these as we are not assuming a reward signal \deleted{and are dependent on fewer known POMDP parameters} during training. 

Vision, memory, and controller modules are combined for learning a model of the world before learning a decision model in Ha and Schmidhuber's World Models \cite{ha2018world}. A predictive model is trained in an unsupervised manner, permitting the agent to learn policies completely within its learned latent space representation of the environment. The main difference is that they first approximate the state distribution using a variational autoencoder, producing the encoded latent space. In contrast, our representation is learned such that it is maximally predictable for the predictor network.

Similar to our training setup, Oh et al.~\cite{oh2015action} predict future frames in ATARI environments conditioned on actions. The predicted frames are used for learning the transition function of the environment, e.g.\ for improving exploration by informing agents of which actions are more likely to result in unseen states. Our work differs as we are acting within a learned latent space and not the full input space \added{and our representations are used in a classical planning paradigm with start and goal states instead of a reinforcement learning one}. 
\subsection*{Visual planning}

We define visual planning as the problem of synthesizing an action sequence to generate a target state from an initial state, and all the states are observed as images.  Variational State Tabulations~\cite{corneil2018efficient} learn a state representation in addition to a transfer function over the latent space. However, their observation space is discretized into a table using a variational approach, as opposed to our continuous representation. \added{A continuous representation circumvents the problem of having to determine the size of such a table in advance or during training.} Similarly, Cuccu et al.~\cite{cuccu2018playing} discretize visual input using unsupervised vector quantization and use that representation for learning controllers for Atari games.

Inspired by classic symbolic planning, Regression Planning Networks \cite{{xu1029regression}} create a plan backward from a symbolic goal. We do not have access to high-level symbolic goal information for our method and we assume that only high-dimensional visual cues are received from the environment.

Topological memories of the environment are built in Semi-parametric Topological Memories  \cite{savinov2018semi} after \replaced{being provided with observation sequences from humans exploring the environment.}{manual exploration.} Nodes are connected if a predictor estimates that they are close. The method has problems with generalization, which are reduced in Hallucinative Topological Memories \cite{liu2020hallucinative}, where the method also admits a description of the environment, such as a map or a layout vector, which the agent can use during planning. Our visual planning method does not receive any additional information on unseen environments and does not depend on manual exploration during training.

Causal InfoGAN \cite{kurutach2018learning} and related methods \cite{wang2019learning} are based on generative adversarial networks (GANs) \cite{goodfellow2014generative}, inspired by InfoGAN in particular \cite{chen2016infogan}, for learning a plannable representation. A GAN is trained for encoding start and goal states and they plan a trajectory in the representation space as well as reconstructing intermediate observations in the plan. \replaced{Our method is different as it does not need to reconstruct the observations and the forward model is directly optimized for prediction.}{Our method is different as it learns the representation as well as the forward model end-to-end, giving up the need for reconstruction.}

\subsection*{Prediction-based Representation learning}

 In Predictable Feature Analysis \cite{richthofer2015predictable}, representations are learned that are predictable by autoregression processes. Our method is more flexible and scales better to higher dimensions as the predictor can be any differentiable function.

Using the output of other networks as prediction targets instead of the original pixels is not new. The case where the output of a larger model is the target for a smaller model is known as knowledge distillation \cite{hinton2015distilling} \cite{bucilua2006model}. This is used for compressing a model ensemble into a single model. Vondrick et al. \cite{vondrick2016anticipating} learn to make high-level semantic predictions of future frames in video data. Given a current frame, a neural network predicts the representation of a future frame. Our approach is not constrained only to pre-trained representations, we learn our representation together with the prediction network. Moreover, we extend this general idea by also admitting an action as the input to our predictor network.

\section*{Materials and methods}

In this work, we study different representations for learning the transition function of a partially observable MDP (POMDP) and propose a network that jointly learns a  representation with a prediction model and apply it for latent space planning. We summarize here the different ingredients of the LARP network -- our proposed solution. More detailed descriptions will follow in later sections.

\textbf{Training the predictor network:} We use a two-stream fully connected neural network (see~\nameref{S2_appendix} for details) to predict the representation of the future state given the current state's representation and the action bridging those two states. The predictor module is trained with a simple mean-squared error term.

\textbf{Handling constant solutions:} The representation could be transferred from other domains or learned from scratch on the task. If the representation is learned simultaneously with an \replaced{estimate}{estimation} of a Markov decision process's (MDP) transition function, precautions must be taken such that the prediction loss is not trivially minimized by a representation that is constant over all states.  We consider three approaches for tackling the problem: sphering the output, regularizing with a contrastive loss term, and regularizing with a reconstructive loss term.

\textbf{Searching in the latent space:} Combining the representation with the predictor network, we can search in the latent space until a node is found that has the largest similarity to the representation of the goal viewpoint using a modified best-first search algorithm.

\textbf{NORB environment:} We use the NORB data set~\added{\cite{lecun2004learning}} for our experiments.  This data set consists of images of  objects
from different viewpoints and we create viewpoint-matching tasks from the data set. 

\subsection*{Partially-observable Markov decision processes}

Many real-world situations can be \replaced{described as}{reduced to} partially-observable Markov decision processes (POMDP) \cite{cassandra1998survey}.  Under this general framework, there are \textit{agents}  taking \textit{actions} $A$ changing their environment's \textit{state} $S$. The information that they have on the states comes from the \textit{observations} $O$ they perceive as not all the details of the environment's state \replaced{are}{is} generally known. How the environment's state $S_t$ at a time step $t$ changes after an action $A_t$ is performed is governed by the transition function $T(S_t, A_t)$. \replaced{Additionally,}{After the action $A_t$ is realized,} the system grants the agent a reward $R_t$, whose value is determined by the reward function $R(S_t, A_t)$.  In this work, we will refer the objective of the agent in the environment, such as reaching a goal location, as its \textit{task}

 The goal of the agent is to maximize the expected cumulative discounted future reward $\mathbb{E} \left[ \sum_{t=0}^\infty \gamma^t R_t \right]$ in the environment, where a discount factor $\gamma \in [0, 1]$ is given in the POMDP. Model-free RL methods learn an optimal policy to maximize the reward without modeling the environment transition function separately. We learn a transition function for a low-dimensional representation, as well as the representation itself, of the observation space, which we use to maximize the reward in the POMDP. Our method is thus a model-based RL method, since we effectively learn a model of the environment's dynamics.
\subsection*{On good representations}
We rely on heuristics to provide sufficient evidence for a good --- albeit not necessarily optimal --- decision at every time step to reach the goal. Here we use the Euclidean distance in representation space: a sequence of actions is preferred if their end location is closest to the goal. The usefulness of this heuristics depends on how well and how coherently the Euclidean distance encodes the actual distance to the goal state in terms of the number of actions.

A learned predictor network approximates the transition function of the environment for planning in the latent space defined by some representation. This raises the question: what is the ideal representation for latent space planning? Our experiments show that an openly available, general-purpose representation, such as a pre-trained VGG16 \cite{simonyan2014very}, can already provide sufficient guidance to apply such heuristics effectively. Better still are representation models that are trained on the data at hand, for example, uniform manifold approximation and projection (UMAP) \cite{mcinnes2018umap} or variational auto-encoders (VAEs) \cite{kingma2013auto}.

One might, however, ask what a particularly suited representation might look like when attainability is ignored. It would need to take the topological structure of the underlying data manifold into account, so that the Euclidean distance becomes a good proxy for the geodesic distance. One class of methods that satisfy this are spectral embeddings, such as Laplacian Eigenmaps (LEMs) \cite{belkin2003laplacian}. Their representations are smooth and discriminative which is ideal for our purpose. However, they do not easily produce out-of-sample embeddings, so they will only be applied in an in-sample fashion to serve as a control experiment\replaced{, yielding optimal performance.}{-- and we get perfect performance with LEMs.}

\subsection*{Predictor network}

As the representation is used by the predictor network, we want it to be predictable. Thus, we optimize the representation learner simultaneously with the predictor network, in an end-to-end manner.

Suppose we have a representation map $\phi$ and a training set of $N$ labeled data tuples
$(X_t = [O_t, A_t], Y_t = O_{t+1})$, where $O_t$ is the observation at time step $t$ and $A_t$ is an action resulting in a state with observation $O_{t+1}$. We train the predictor $f$, parameterized by $\theta$, by minimizing the mean-squared error\replaced{:}{loss over $f$'s parameters:}
\begin{equation}
\underset{\theta}{\text{argmin}} \ \mathcal{L}_{\text{prediction}}(\added{\mathcal{D}}, \theta) = \underset{\theta}{\text{argmin}} \  \frac{1}{N} \sum_{t=1}^N \big \Vert \phi(O_{t+1}) - f_\theta(\phi(O_t), A_t) \big \Vert ^2    
\end{equation}
\noindent where $\added{\mathcal{D}} = \{ (X_t, Y_t) \}_{t=0}^N $ is our set of training data.

We construct $f$ as a two-stream, fully connected, neural network (\nameref{S2_appendix}). Using this predictor we can carry out planning in the latent space defined by $\phi$. By planning, we mean that there is a start state with observation $O_{\text{start}}$ and a goal state with $O_{\text{goal}}$ and we want to find a sequence of actions connecting them.

The network outputs the expected representation after acting. Using this, we can formulate planning as a classical pathfinding or graph traversal problem.

\subsection*{Avoiding trivial solutions}

In the case where $\phi$ is trainable and parameterized by $\eta$, the loss for the whole system that only cares about maximizing predictability is
\begin{equation}
\underset{\theta, \eta}{\text{argmin}} \ \mathcal{L}_{\text{prediction}}(\added{\mathcal{D}}, \theta, \eta) = \underset{\theta, \eta}{\text{argmin}} \  \frac{1}{N} \sum_{t=1}^N \left(\phi_\eta(O_{t+1}) - f_\theta(\phi_\eta(O_t), A_t) \right)^2
\label{uselessloss}
\end{equation}
for a given data set \replaced{$\mathcal{D}$.}{$D$.} With no constraints on the family of functions that $\phi$ can belong to, we run the risk that the representation collapses to a constant. Constant functions $\phi = c$ trivially yield zero loss for any set \replaced{$\mathcal{D}$}{$D$} if $f_\theta$ outputs the input state again for any $A$, i.e $f(\phi(\cdot), A) = \phi(\cdot)$:

Constant representations are optimal with respect to predictability but they are unfortunately useless for planning, as we need to discriminate different states. This objective is not present in the proposed loss function in Eq.~\eqref{uselessloss} and we must thus add a constraint or another loss term to facilitate differentiating the different states.

There are several ways to limit the function space such that constant functions are not included, for example with decoder \cite{goroshin2015learning} or adversarial \cite{denton2017unsupervised} loss terms. In this work, we do this with \textbf{(i)} a sphering layer, \textbf{(ii)} a contrastive loss, or \textbf{(iii)} a reconstructive loss.

\subsubsection*{(i) Sphering the output} The problem of trivial solutions is solved in Slow Feature Analysis \cite{wiskott2002slow} and related methods \cite{schuler2018gradient} \cite{escalante2013solve} by constraining the overall covariance of the output to be $I$. Including this constraint to our setting yields the optimization formulation:
\begin{equation}
\begin{aligned}
& \underset{\eta, \theta}{\text{minimize}}
& & \mathcal{L}_{\text{prediction}}(\added{\mathcal{D}}, \theta, \eta) \\
& \text{subject to}
& & \mathbb{E}_{\added{\mathcal{D}}}\left[ \phi_\eta  \right] \hspace{7pt}= 0 \hspace{47pt}\hspace{0.1em}\text{(zero mean)} \\
&&&  \mathbb{E}_{\added{\mathcal{D}}}\left[ \phi_\eta\phi_\eta^T \right]= I\, \ \ \   \ \ \  \ \ \ \ \ \,\text{(unit covariance)} \\
\end{aligned}
\end{equation}
\noindent We enforce this constraint in our network via architecture design. The last layer performs differentiable sphering \cite{schuler2018gradient}~\cite{10.1007/978-3-030-30179-8_15} of the second-to-last layer's output using the whitening matrix $\bm{W}$. We get $\bm{W}$ using power iteration of the following iterative formula:
\begin{equation}
\bm{u}^{[i+1]} = \frac{\bm{T}\bm{u}^{[i]}}{|| \bm{T}\bm{u}^{[i]} ||}
\end{equation}
\noindent 

\begin{figure}[htb]
\begin{adjustwidth}{-0.63in}{0in}
\caption{{\bf Predictive representation learning with sphering regularization.} The observations $O_t$, and the resulting observation $O_{t+1}$ after the action $A$ has been performed in $O_t$, are passed through the representation map $\phi$, whose outputs are passed to a differentiable sphering layer before being passed to $f$. The predictive network $f$ minimizes the loss function $\mathcal{L}$, which is the mean-squared error between $ \phi(O_t) = \rho_t$ and $ \phi(O_{t+1}) = \rho_{t+1}$.}
\centering
\includegraphics[width=1.0\textwidth]{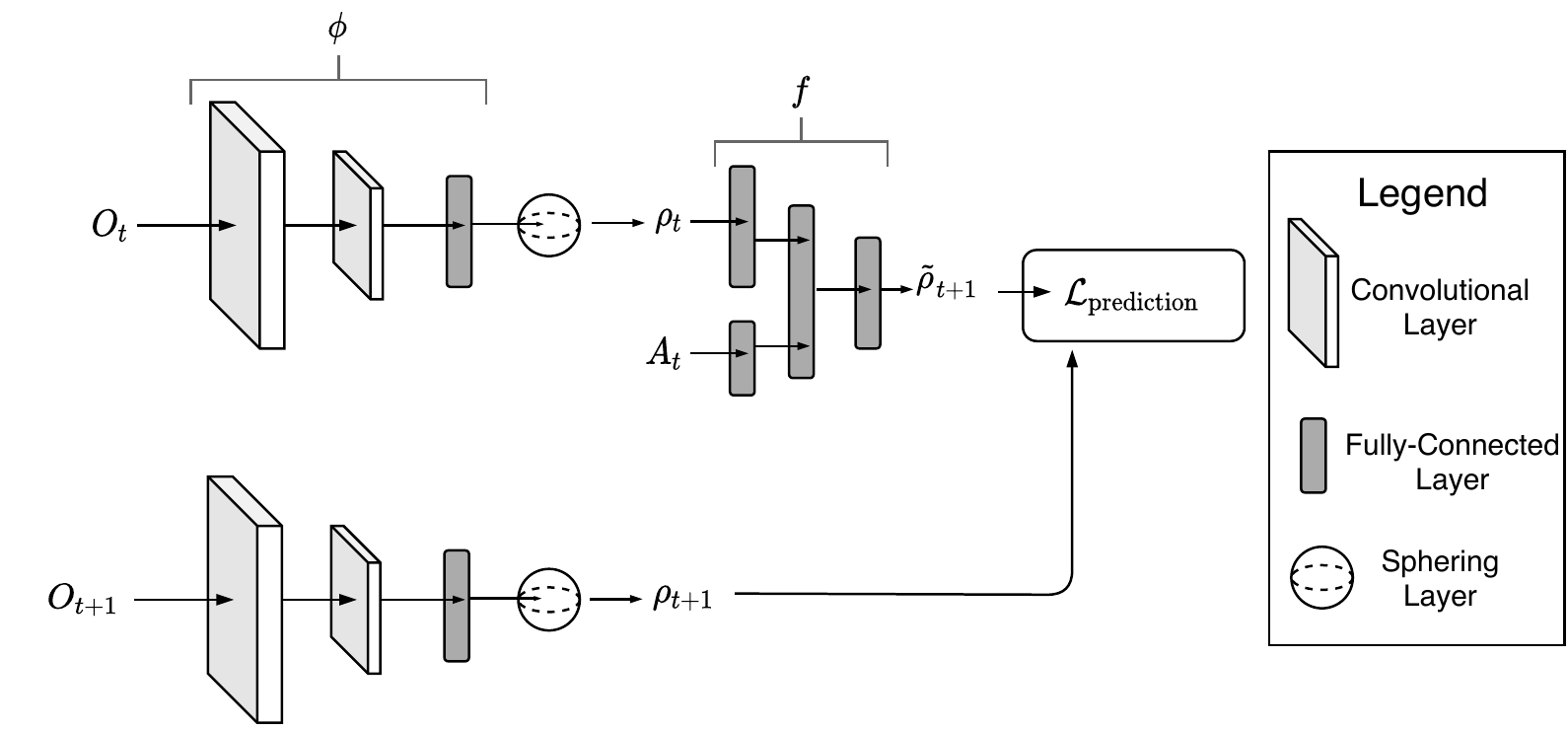}
\label{spheringdiagram}
\end{adjustwidth}

\end{figure}

\noindent \added{where the superscript $i$ tracks the iteration number and $\bm{u}^{[0]}$ can be an arbitrary vector.} \replaced{The power iteration algorithm}{Which}
converges to the largest eigenvector  $\bm{u}$ of a matrix $\bm{T}$ in a few hundred, quick iterations. The eigenvalue $\lambda$ is determined and we subtract the eigenvector from the matrix:
\begin{equation}
\bm{T} \leftarrow \bm{T} - \lambda \bm{u} \bm{u}^T
\end{equation}
\noindent the process is repeated until the sphering matrix is found
\begin{equation}\bm{W} = \sum_{j=0} \frac{1}{\sqrt{\lambda_j}}\bm{u}_j \bm{u}_j^T
\end{equation}
\noindent The whole system, including the sphering layer, can be seen in Fig~\ref{spheringdiagram}, \added{with an abstract convolutional neural network as the representation $\phi$ and a fully-connected neural network as the prediction function $f$}.

\subsubsection*{(ii) Contrastive loss} Constant solutions can also be dealt with in the loss function instead of via architecture design. Hadsell et al. \cite{hadsell2006dimensionality} propose to solve this with a loss function that pulls together the representation of similar objects (in our case, states that are reachable from each other with a single action) but pushes apart the representation of dissimilar ones:
\begin{equation}
L_{\text{contrastive}}{(O, O^{'} )} = \begin{cases} || \phi (O) - \phi(O^{'})  || & \text{if } O, O^{'} \text{ are similar} \\
                      \max(0, M - || \phi (O) - \phi(O^{'})  || )          & \text{otherwise} \end{cases}
\end{equation}
\noindent where $M$ is a margin and $|| \cdot||$ is some --- usually the Euclidean --- norm. The representation of dissimilar objects is pushed apart only if the inequality
\begin{equation}
|| \phi (O) - \phi(O^{'})  || < M
\end{equation}
is violated. During each training step, we compare each observation to a similar and a dissimilar observation simultaneously~\cite{schroff2015facenet} by passing a triplet of (positive, anchor, negative) observations during training to three copies of $\phi$. In our experiments, the positive corresponds to the predicted embedding of $O_{t+1}$ given $O_t$ and $A_t$,  the anchor is the true resulting embedding after an action $A_t$ is performed in state  $O_t$ and the negative \added{$\phi(O_n)$} is \replaced{the representation of an arbitrarily chosen observation}{any other embedding} that is not reachable from $\phi(O_t)$ with a single action. \added{For environments where this is determinable, such as in our experiments, this can be assessed from the environment's full state. When this information isn't available, ensuring for $\phi(O_n)$ and $\phi(O_t)$ that $|n-t|>2$ is a good proxy, even though this can result in some incorrect triplets. For example, when the agent runs in a self-intersecting path.}

We define the representation of the observation at time step $t$ as $\rho_t = \phi(O_t)$ and the next-step prediction $\Tilde{\rho}_{t+1} := f\left(\phi(O_t), A_t)\right)$ for readability \added{and our (positive, anchor, negative) triplet is thus $\left(\Tilde{\rho}_{t+1}, \phi(O_{t+1}), \phi(O_n) \right)$} and \added{we} minimize \added{the triplet loss}:
\begin{equation}
\mathcal{L}_{\text{contrastive}}(O_t, O_{t+1}, O_n, A_t) = || \rho_{t+1} - \Tilde{\rho}_{t+1}  || + \max(0, M - || \rho_{t+1} - \rho_n  || )  
\label{ourcontrastive}
\end{equation}

\noindent It would seem that $\rho _{t+1}$ and $\Tilde{\rho}_{t+1}$ are interchangeable since the second term is included only to prevent the representation from collapsing into a constant. However, if the loss function is
\begin{equation}
\mathcal{L}_{\text{contrastive}}(O_t, O_{t+1}, O_n, A_t) = || \rho_{t+1} - \Tilde{\rho}_{t+1}  || + \max(0, M - ||  \Tilde{\rho}_{t+1} - \rho_n  || )  
\label{badcontrastive}
\end{equation}
then the network is rewarded during training for making $f$ poor at predicting the next representation instead of simply pushing the representation of $O_{t}$ and $O_n$ away from each other.

\begin{figure}[htb]
\caption{ {\bf Predictive representation learning with contrastive loss regularization.} We minimize the contrastive loss function $\mathcal{L}_{\text{contrastive}}$ (Eq. \ref{ourcontrastive}). The predicted future representation $\Tilde{\rho}_{t+1}$ is pulled toward the next step's representation $\rho_{t+1}$. However, $\rho_{t+1}$ is pushed away from the negative state's representation $\rho_{n}$ if the distance between them is less than $M$. The observation $O_n$ is randomly selected from those that are not reachable from $O_{t}$ with a single action.}
\centering
\includegraphics[scale=.99]{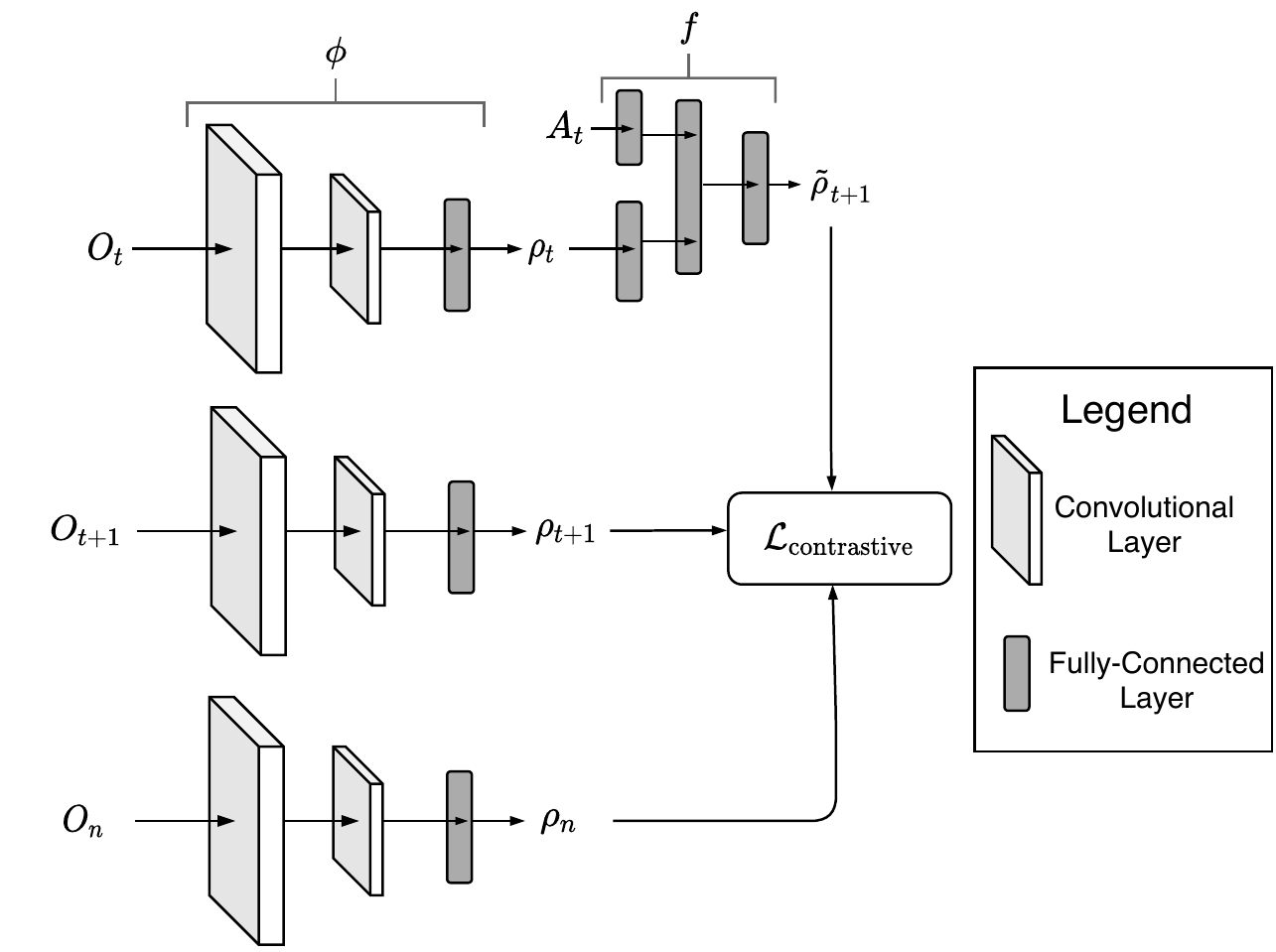} 
\label{contrastivefigure}
\end{figure}

There are two main ways to set the margin $M$, one is dynamically determining it per batch \cite{sun2014deep}. The other, which we choose, is constraining the representation to be on a hypersphere using $L_2$ normalization and setting a small constant margin such as $m = 0.2$ \cite{schroff2015facenet}. The architecture for the training scheme using the contrastive loss regularization is depicted in Fig.~\ref{contrastivefigure}.

\subsubsection*{(iii) Reconstructive loss} Trivial solutions are avoided by Goroshin et al. \cite{goroshin2015learning} by introducing a decoder network $D$ to a system that would otherwise converge to a constant representation. We incorporate this intuition into our framework with the loss function

\begin{align}
\label{decoderloss}
     \mathcal{L}_{\text{decoder}} (O_{t}, O_{t+1}, A_t) = \mathcal{L}_{\text{prediction}}(O_{t},O_{t+1}) + \mathcal{L}_{\text{reconstruction}}(O_{t},O_{t+1}) \nonumber   && \\
    =\left( \rho_t - \Tilde{\rho}_{t+1}\right)^2 + \alpha  \left(O_{t+1}  - D \left( \Tilde{\rho}_{t+1} \right) \right) ^2 \hspace{12pt} && \text{}
\end{align}

\begin{figure}
\caption{{\bf Predictive representation learning with decoder loss regularization.} \replaced{At time step $t$, the observation $O_t$ is passed to the representation $\phi$. This produces $\rho_t$ which is passed, along with the action $A_t$ at time step $t$, to the predictor network $f$. This produces the predicted $\Tilde{\rho}_{t+1}$ which is compared to $\rho_{t+1} = \phi(O_{t+1})$ in the mean-squared error term $\mathcal{L}_\text{prediction}$. The prediction $\Tilde{\rho}_{t}$ is also passed to the decoder network $D$. We then compare $\Tilde{O}_{t+1} = D(\Tilde{\rho}_{t+1})$ with $O_{t+1}$ in $\mathcal{L}_\text{decoder}$, another mean-squared loss term. The final loss is the sum of these two loss terms $\mathcal{L}_\text{total} = \mathcal{L}_\text{prediction}+\mathcal{L}_\text{decoder}$.}{ The $\mathcal{L}_\text{prediction}$ and $\mathcal{L}_\text{decoder}$ boxes both return the mean-squared error of their inputs and the full loss  $\mathcal{L}_\text{total}$ is the sum of those two terms.}}
\centering
\includegraphics[scale=.99]{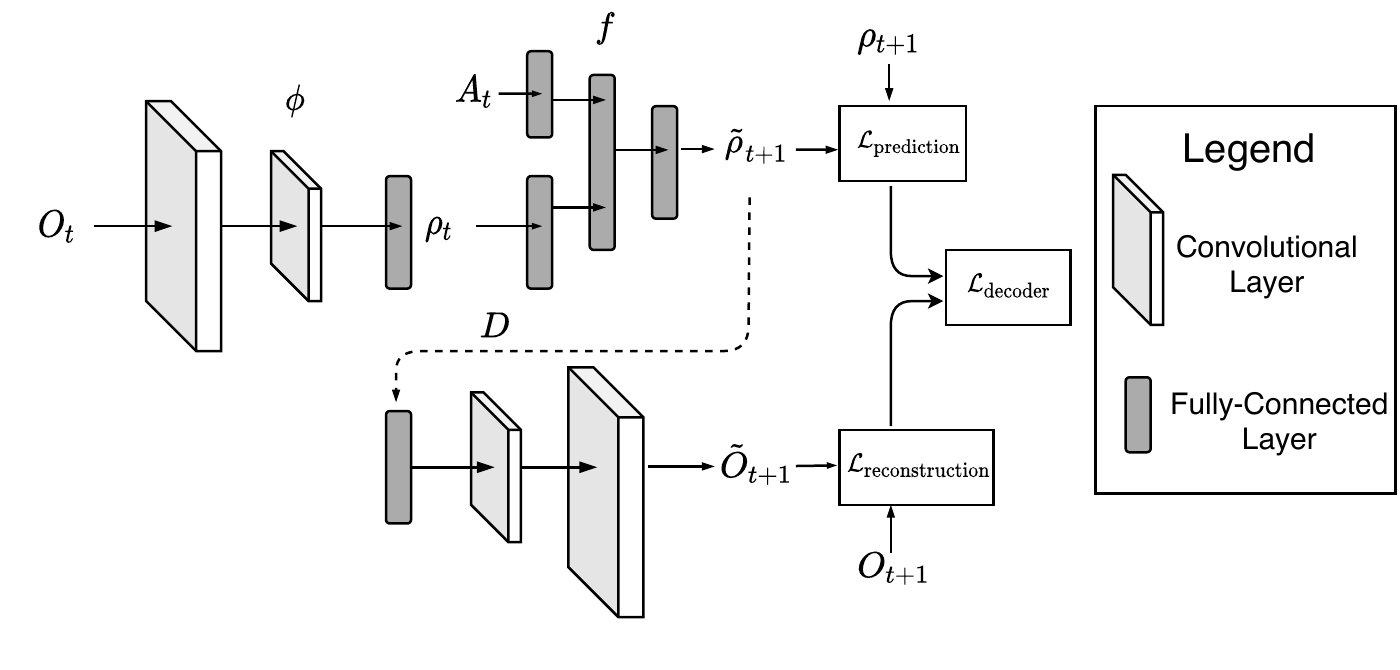} 
\label{decoderlossgraphic}
\end{figure}

\noindent \added{where $\alpha$ is a positive, real coefficient to control the regularization strength.} \added{Fig.~\ref{decoderlossgraphic} shows how the models and loss functions are related during the training of the representation and predictor using both a predictive and a reconstructive loss term.}

The desired effect of the regularization can also be achieved by replacing the second term in Eq.~\eqref{decoderloss} with $\alpha  \left( O_{t+1}  - D \left( \rho_{t+1}\right) \right) ^2$. By doing this we would maximize the reconstructive property of the latent code in and of itself, which is not inherently useful for planning. We instead add an additional level of predictive power in $f$: in addition to predicting the next representation, its prediction must also be useful in conjunction with the decoder $D$ for reconstructing the new true observation. 

This approach can have the largest computational overhead of the three, depending on the size of the decoder. We construct the decoder network $D$ such that it closely mirrors the architecture of $\phi$, with convolutions replaced by transposed convolutions and max-pooling replaced by upsampling.

\subsection*{Training the predictor network}

We train the representation network and predictor network jointly by minimizing Eq.~(2), Eq.~(10) or Eq.~(12). The predictor network can also be trained on its own for a fixed representation map $\phi$.  In this case, $f$ is tasked as before with predicting $\phi(O_{t+1})$ after the action \replaced{$A_t$}{$A$} is performed in the state with observation $O_t$ by minimizing the mean-squared error between $f(\phi(O_{t}), A_t)$ and $\phi(O_{t+1})$. The networks are built with Keras \cite{chollet2015keras} and optimized with rmsprop \cite{tieleman2012lecture}. 

\subsection*{Planning in transition-learned domain representation space}
We use a modified best-first search algorithm with the trained representations for our experiments (Algorithm \ref{algo: gs_algo}). From a given state, the agent performs a simulated rollout to search for the goal state. For each action, the initial observation is passed to the predictor function along with the action. This results in a predicted next-step representation, which is added to a \replaced{set}{queue}. The actions taken so far and resulting in each prediction are noted also. The representation that is closest to the goal (using for example the Euclidean distance) is then taken for consideration and removed from the \replaced{set}{queue}. This process is repeated until the maximum number of trials is reached. The algorithm then outputs the sequence of actions resulting in the predicted representation that is the closest to the goal representation.

\begin{algorithm}
\caption{Perform a simulated rollout to find a state that is maximally similar to a goal state. Output a \replaced{sequence}{plan} to reach the found state from the start state.}\label{algo: gs_algo}
\begin{algorithmic}[1]
\REQUIRE $O_{\text{start}}$, $O_{\text{goal}}$, max trials $m$, action set $\mathcal{A}$,  representation map $\phi$ and predictor function $f$ 
\ENSURE A \replaced{sequence}{plan} of actions $(A_0, \dots, A_n)$ connecting the start state to the goal state
\STATE Initialize the \replaced{set}{queue}  $Q$ \added{of unchecked representations} with the representation of the start state $\phi(O_{\text{start}})$

\STATE \added{ Initialize the dictionary $P$ of representation-path pairs  with the initial representation mapped to an empty sequence: $P[\phi(O_{\text{start}})]\leftarrow(\varnothing$)
}

\STATE Initialize \replaced{the empty set of checked representations $C \leftarrow \varnothing$}{the set of checked states D as empty}

\FOR{$k \leftarrow 0 \text{ to } m$}

    \STATE Choose $\rho'$ \added{$\leftarrow$}   $\underset{\rho \in Q}{\text{argmin}} ||\rho - \phi(O_{goal}) ||$
    \STATE Remove $\rho'$ from $Q$ and add it to $\added{C}$
    \FORALL{actions $A \in \mathcal{A}$} 
        \STATE Get a new estimated representation $\rho^* \added{\leftarrow} f(\rho', A)$ and add it to the \replaced{set}{queue} Q
        \STATE \added{Concatenate $A$ to the end of $P[p']$ and associate the resulting sequence with $\rho^*$ in the dictionary:  $P[\rho^*] \leftarrow P[\rho'] ^\frown  (A)$}
        \STATE \deleted{Associate with $\rho^*$ the sequence $P^*$ of actions to reach it from the start state}
    \ENDFOR
\ENDFOR
\STATE \added{Find the most similar representation to the goal: $\rho_{\text{result}} \leftarrow  \underset{\rho \in Q \cup \added{C}}{\text{argmin}} ||\rho - \phi(O_{goal}) ||$} 
\RETURN{the sequence \replaced{$P[\rho_{\text{result}}]$}{$P_{\text{out}}$ that is associated with $\rho_{\text{out}} $=\\$  \underset{\rho \in Q \cup D}{\text{argmin}} ||\rho - \phi(O_{goal}) ||$}}
\end{algorithmic}
\end{algorithm}

To make the algorithm faster, we only consider paths that do not take us to a state that has already been evaluated,
even if there might be a difference in the predictions from going this roundabout way. That is, if a permutation of the actions in the next path to be considered is already in an evaluated path, it will be skipped. This has the same effect as transposition tables used to speed up search in game trees. Paths might be produced with redundancies, which can be amended with path-simplifying routines (e.g. take one step forward instead of one step left, one forward then one right).

We do Model-Predictive Control \cite{garcia1989model}, that is, after a path is found, one action is performed and a new path is recalculated, starting from the new position. Since the planning is possibly over a long time horizon, we might have a case where a previous state is revisited. To avoid loops resulting from this, we keep track of visited state-action pairs and avoid an already chosen action for a given state.

\subsection*{NORB viewpoint-matching experiments}

    For our experiments, we create an OpenAI Gym environment based on the small NORB data set \cite{lecun2004learning}. The code for the environment is available at \url{https://github.com/wiskott-lab/gym-norb} and requires the pickled NORB data set hosted at \url{https://s3.amazonaws.com/unsupervised-exercises/norb.p}. The data set contains 50 toys, each belonging to one of five categories: four-legged animals, human figures, airplanes, trucks, and cars. Each object has stereoscopic images under six lighting conditions, 9 elevations, and 18 azimuths (in-scene rotation). In all of the experiments, we train the methods on nine car class toys, testing on the other toys.

Each trial in the corresponding RL environment revolves around a single object under a given lighting condition. The agent is presented with a start and a goal viewpoint of the object and transitions between images until the current viewpoint matches the goal where each action operates the camera. To be concrete, the actions correspond to turning a turntable back and forth by $20^{\circ}$, moving the camera up or down by $5^{\circ}$ and, in one experiment, changing the lighting. The trial is a success if the agent manages to change viewpoints from the start position until the goal viewpoint is matched in fewer than twice the minimum number of actions necessary.

We compare the representations learned using the three variants of our method to five representations from the literature, namely  \textbf{(i)} Laplacian Eigenmaps  \cite{belkin2003laplacian}, \textbf{(ii)} the second-to-last
layer of VGG16 pre-trained on ImageNet\cite{deng2009imagenet}, \textbf{(iii)} UMAP embeddings \cite{mcinnes2018umap}, \textbf{(iv)} convolutional encoder \cite{masci2011stacked} and \textbf{(v)} VAE codes \cite{kingma2013auto}. As fixed representations do not change throughout the training, they can be saved to disk, speeding up the training. As a reference, we consider three reinforcement learning methods working directly on the input images: \textbf{(i)} a Deep Q-Networks (DQN)~\cite{mnih2013playing}, \textbf{(ii)} Proximal Policy Optimization (PPO)~\cite{schulman2017proximal} and \textbf{(iii)} World Models~\cite{ha2018world}.

The data set is turned into a graph for search by setting each image as a node and each viewpoint-changing action as an edge. The task of the agent is to transition between neighboring viewing angles until a goal viewpoint is reached. The total number of training samples is fixed at 25600. For our method, a sample is a single ($O_t$, $O_{t+1}$, $A_t$) triplet to be predicted while for the regular RL methods it is a ($O_t$, $O_{t+1}$, $A_t$, $\rho_t$) tuple.

\subsection*{Model \replaced{architectures}{Architectures}}

\subsubsection*{Input} The network $\phi$ encodes the full NORB input, a $96 \times 96$ pixel grayscale image, to lower-dimensional representations. The system as a whole receives the image from the current viewpoint, the image of the goal viewpoint and a one-hot encoding of the taken action. The image inputs are converted from integers ranging from 0 to 255 to floating point numbers ranging from 0 to 1.

\subsubsection*{Representation \replaced{learner $\phi$ architecture}{Learner $\phi$ Architecture}} We use the same architecture for the $\phi$ network in all of our experiments except for varying the output dimension,  Table~\ref{table:phiarchiteture}. 

\begin{table}[htb]\begin{center}
  \captionof{table}{{\bf Representation network architecture.}}
  \label{table:phiarchiteture}
 \begin{tabular}{l r r r l r} 
 Layer &   Filters / Units  & Kernel \replaced{size}{Size} & Strides & Output shape & Activation   \\ [0.5ex] 
\arrayrulecolor{gray} \hline \hline
     Input  &  &  &  & (96, 96, 1) &  \\ 
     \arrayrulecolor{lightgray} \cline{1-6} 
     Convolutional  & 64 & $5 \times 5$ & $2 \times 2$ & (45, 45, 64) & ReLU \\ 
     \arrayrulecolor{lightgray} \cline{1-6} 
     \cline{1-5} Max-pooling &  & & $2 \times 2$  & (22, 22, 64) &  \\ 
   \cline{1-6}  Convolutional & 128 & $5 \times 5$ & $2 \times 2$ & (9, 9, 128) & ReLU \\ 
     \cline{1-6} Flatten &  &  & & (10368) & \\ 
     \cline{1-6} Dense & 600 & && (600) & ReLU \\
     \cline{1-6} Dense & \#Features& && (\# Features) & Linear  \\
\end{tabular}
\end{center}\end{table} 

\subsubsection*{Regularizing \replaced{decoder architecture $D$}{Decoder Architecture $D$}} The decoder network $D$ has the architecture listed in Table~\ref{table:decoder}. It is designed to approximately inverse each operation in the original $\phi$ network.

\begin{adjustwidth}{-3.25in}{0in}
\begin{table}[htb]
\begin{center}
  \captionof{table}{{\bf Regularizing decoder architecture.} The upsampling layer uses linear interpolation, BN stands for Batch Normalization and CT stands for Convolutional Transpose. }
  \label{table:decoder}
 \begin{tabular}{l r r r r l } 
 Layer &  Filters / Units & Kernel Size & Strides & Output shape & Activation  \\ [0.5ex] 
\arrayrulecolor{gray} \hline \hline
     Input &   & &&(\# Features) & \\
    \arrayrulecolor{lightgray} \cline{1-6}
     Dense & 512 & &&(512) & ReLU\\
    \arrayrulecolor{lightgray} \cline{1-6}BN &  & & & (512) &   \\ 
    \cline{1-6} Dense & 12800 & &&(12800) &ReLU\\
    \cline{1-6} BN &  & & & (12800) &  \\ 
    \cline{1-6} Reshape &  & &   & (10, 10, 128) & \\ 
   \cline{1-6}  CT & 128 & $5 \times 5$ & $2 \times 2$ & (23, 23, 128) & ReLU \\ 
   \cline{1-6} Upsampling &  & &  $2 \times 2$  & (46, 46, 128) &\\ 
   \cline{1-6} BN &  & &&(46, 46, 128)    &\\ 
    \cline{1-6} CT  & 64 & $5 \times 5$ & $2 \times 2$ & (95, 95, 64) & ReLU\\ 
    \cline{1-6} BN &  & &  & (95, 95, 64) & \\ 
    \cline{1-6} CT & 1 & $2 \times 2$ & $1 \times 1$ & (96, 96, 1) & Sigmoid \\ 
\end{tabular}
\end{center}
\end{table} 
\end{adjustwidth}

\subsubsection*{Predictor \replaced{network}{Network}  $f$}  The predictor network $f$ is a two-stream dense neural network. Each stream consists of a dense layer \added{with a rectified linear unit (ReLU) activation} followed by a batch normalization \added{(BatchNorm)} layer. The outputs of these streams are then concatenated and passed through 3 dense layers \added{with ReLU activations}, each one followed by a BatchNorm, and then an output dense layer, see Table \ref{table:predictornet}. \deleted{Every dense layer, except the last, is followed by a rectified linear unit activation.}

\begin{adjustwidth}{-3.25in}{0in}
\begin{table}[htb]
\begin{center}
  \captionof{table}{{\bf Represention predictor architecture.} The $\phi$ stream receives the representation as input and the $A$ stream receives the one-hot action as input. Both streams are processed in parallel and then concatened, with each operation applied from top to bottom sequentially. The number of hidden units in the last layer depends on the chosen dimensionality of the representation.}
  \label{table:predictornet}
 \begin{tabular}{l r r l} 
 Layer &  Filters / Units & Output shape & Activation  \\ [0.5ex] 
\arrayrulecolor{gray} \hline \hline
     \ \ \ \ $\phi$ Stream: Input  & & (\# Features) & ReLU\\
\arrayrulecolor{lightgray} \cline{1-4}
     \ \ \ \  $\phi$ Stream: Dense & 256 &  (256) & ReLU\\
    \arrayrulecolor{lightgray} \cline{1-4}\ \ \ \  $\phi$ Stream: Batch Normalization &  &  (256) &    \\ 
    \cline{1-4}\ \  $A$ Stream: Input &  & (\# Actions) & ReLU\\

    \cline{1-4}\ \  $A$ Stream: Dense & 128  &(128)& ReLU\\
    \cline{1-4}\ \  $A$ Stream: Batch Normalization &   & (128) &    \\ 
    \cline{1-4} Concatenate $\phi$ and $A$ streams &   &  (384)  & \\ 

   \cline{1-4}  Dense & 256 &   (256)  & ReLU \\ 
       \cline{1-4} Batch Normalization &  & (256) &    \\

  \cline{1-4}  Dense & 256 &   (256)  & ReLU \\ 
       \cline{1-4} Batch Normalization &   & (256) &    \\
         \cline{1-4}  Dense & 128 &   (128)  & ReLU \\ 

    \cline{1-4} Batch Normalization &   &  (128) &  \\ 
    \cline{1-4}Dense & \# Features &   (\# Features)  & Linear \\ 
\end{tabular}
\end{center}
\end{table} 
\end{adjustwidth}

\section*{Results}
With our empirical evaluation we aim to answer the following research questions:
\begin{enumerate}
\item (Monotonicity) Is the Euclidean distance between a
\added{suitable} representation and the goal representation \replaced{proportional}{similar} to the number of actions that separate them?
\item (Trained predictability) Is training a representation for predictability\added{, as proposed,} feasible?
\item (Dimensionality) What is the best dimensionality of the latent space for our planning tasks?
\item (Solution constraints) In terms of planning performance, what are \replaced{promising}{the best} constraints to place on the representation to avoid trivial solutions?
\item (Benchmarking) How does planning with LARP compare to other methods from the RL literature?

\item (Generalization) \replaced{How well does our method generalize to unseen environments?}{How is the performance of our method affected when we place obstacles in trained environments?}
\deleted{\\ 7. How does our method generalize to different environments?} 
\end{enumerate}
we will refer to these research questions by number below as they get addressed. 

\subsection*{Latent space visualization}

When the representation and predictor networks are trained, we apply Algorithm \ref{algo: gs_algo} to the viewpoint-matching task. As described above, the goal is to find a sequence of actions that connects the start state to the goal state, where the two states differ in their configurations.

To support the qualitative analysis of the latent space, we plot heatmaps of similarity between the goal representation and the predicted representation of nodes during search (Fig~\ref{fig:lem_matching}). Of the 10 car toys in the NORB data set, we randomly chose 9 for our training set and test on the remaining one.

\subsubsection*{In-sample embedding: Laplacian Eigenmaps}
\label{sec:in_sample}
First, we consider research question 1 (monotonicity). In order to get the best-case representation, we embed the toy using Laplacian Eigenmaps.  Embedding a single toy in three dimensions using Laplacian Eigenmaps results in a tube-like embedding that encodes both elevation and azimuth angles, see Fig \ref{fig:azimuth_cylinder}. Three dimensions are needed so that the cyclic azimuth can be embedded correctly as $\sin(\theta)$ and $\cos(\theta)$.

{
\captionsetup{aboveskip=-13pt}
\begin{figure}
\begin{adjustwidth}{-1.65in}{0in}
\centering
\subcaptionbox*{}{\includegraphics[width=0.3\linewidth]{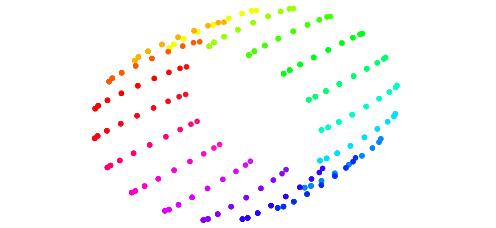}}
\hspace{-1.cm}
\subcaptionbox*{}{\includegraphics[width=0.3\linewidth]{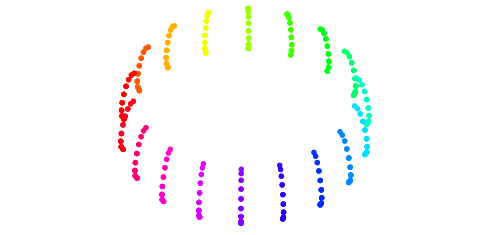}}
\hspace{-1.cm}
\vspace{-0.45cm}
\subcaptionbox*{}{\includegraphics[width=0.3\linewidth]{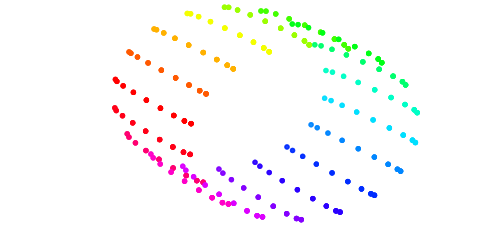}}
\vspace{5mm}
\subcaptionbox*{}{\includegraphics[width=0.3\linewidth]{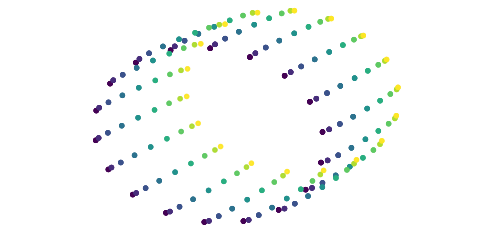}}
\hspace{-1.cm}
\subcaptionbox*{}{\includegraphics[width=0.3\linewidth]{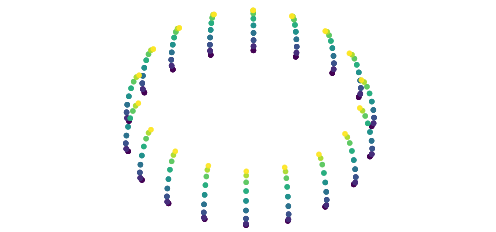}}
\hspace{-1.cm}
\subcaptionbox*{}{\includegraphics[width=0.3\linewidth]{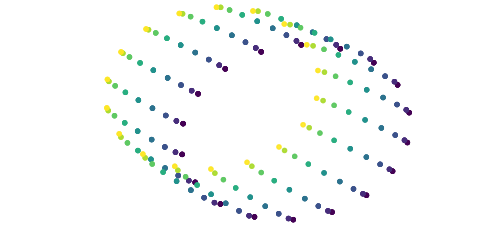}}
\vspace{5mm}
\caption{{\bf Laplacian Eigenmap representation space of a NORB toy.} The three-dimensional Laplacian Eigenmaps of a toy car where elements with the same azimuth values have the same color (top) and where elements with the same elevation values have the same color (bottom). Euclidean distance is a good proxy for geodesic distance in this case.}
\label{fig:azimuth_cylinder}
\end{adjustwidth}
\end{figure}

}

\begin{figure}[htb]
\begin{adjustwidth}{-2.25in}{0in}
\centering
 \captionsetup{width=1.0\linewidth}
  \resizebox*{1.25   \textwidth}{!}{\includegraphics
{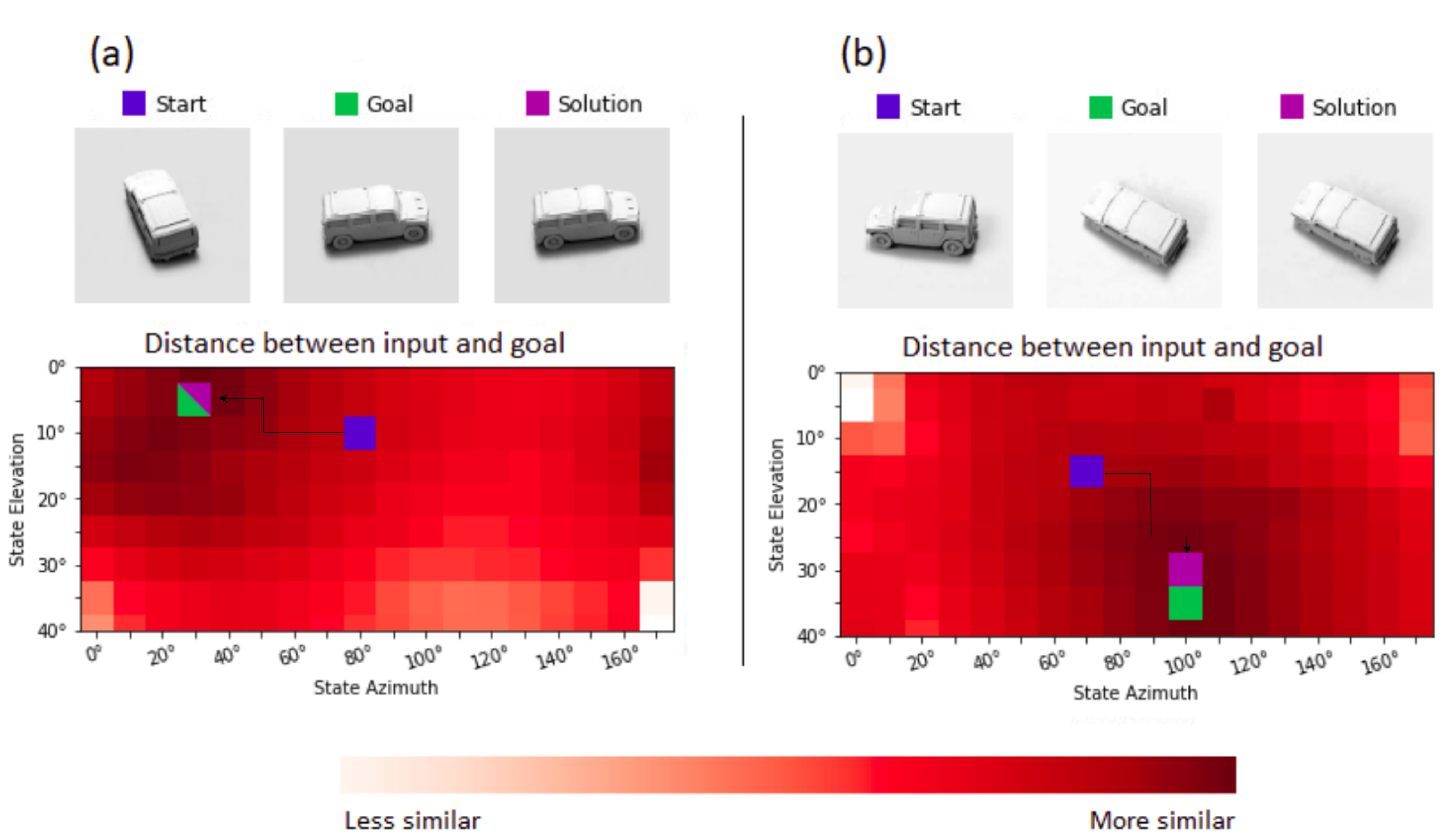}}
\caption{ 
{\bf Heatmap of Laplacian Eigenmap latent space similarity.} Each pixel displays the difference between the predicted representation and the goal representation. Only the start and goal observations are given. The blue dot shows the start state, green the goal, and purple the solution state found by the algorithm.
The search algorithm can rely on an almost monotonically decreasing Euclidean distance between each state's predicted representation and the goal's representation to guide its search.
} 
\label{fig:lem_matching}
\end{adjustwidth}
\end{figure}

If the representation is now used to train the predictor, one would expect that the representation becomes monotonically more similar to the goal representation as the state moves toward the goal. In Fig \ref{fig:lem_matching} we see that this is the case and that this behavior can be effectively used for a greedy heuristics. While the monotonicity is not always exact due to errors in the prediction, Fig \ref{fig:lem_matching} still qualitatively illustrates a best-case scenario.

We conclude from this that \deleted{training a representation for predictability is feasible and that} the Euclidean distance between a current representation and the goal representation is monotonically increasing as a function of the number of actions that separate them. \added{This supports the use of a prediction-based latent space search for planning.}

\subsubsection*{Out-of-sample embedding: pre-trained VGG16 representation}
\begin{figure}
\begin{adjustwidth}{-2.05in}{0in}
\centering
 \captionsetup{width=1.0\linewidth}
  \resizebox*{1.25   \textwidth}{!}{\includegraphics
{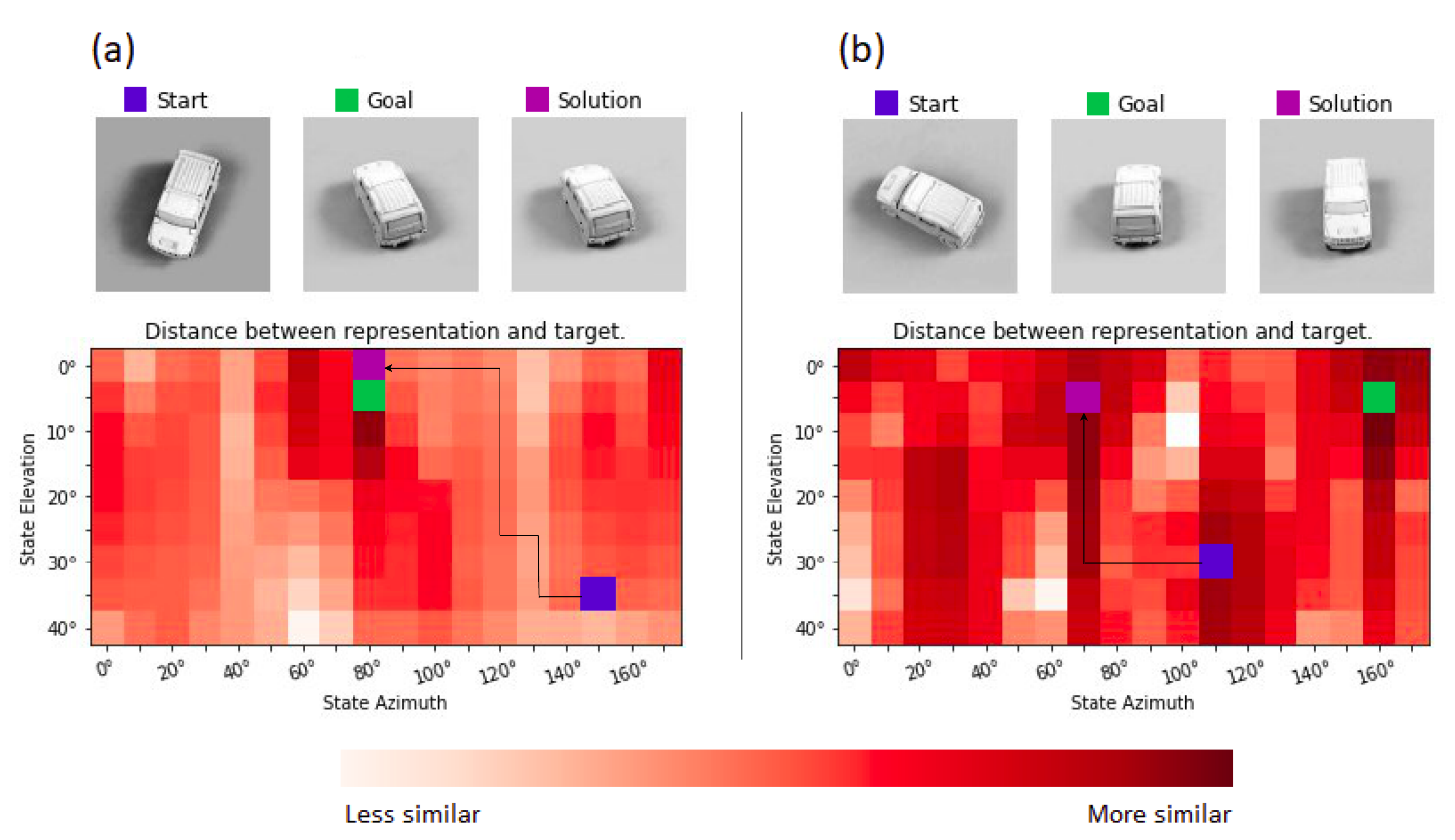}}
\caption{
{\bf Heatmap of VGG16 latent space similarity.}  The predictor network estimates the VGG16 representation of the resulting states as the object is manipulated.
 \textbf{(a)} The goal lies on a hill containing a maximum of representational similarity. \textbf{(b)} The accumulated errors of iterated estimations cause the algorithm to plan a path to a wrong state with a similar shape.
}
\label{fig:allinone}
\end{adjustwidth}
\end{figure}

Next, we consider the pre-trained representation of the VGG16 network to get a representation that generalizes to new objects. We train the predictor network and plot the heat map of the predicted similarity between each state and the goal state, beginning from the start state, in  Fig~\ref{fig:allinone}.

\begin{figure} 
\begin{adjustwidth}{-2.25in}{0in}
\centering
 \captionsetup{width=.9\linewidth}
  \resizebox*{1.25   \textwidth}{!}{\includegraphics
{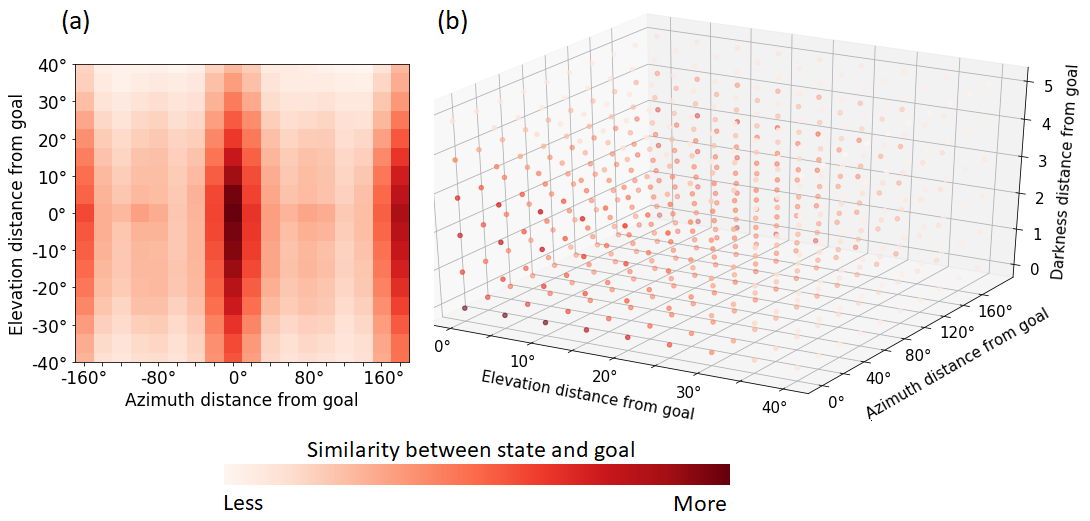}}
\caption{ {\bf Aggregate heat maps of VGG16 representation similarities on test data.} The data is collected as the state space is searched for a matching viewpoint. The pixels are arranged according to their elevation and azimuth difference from the goal state at $(0^{\circ}, 0^{\circ})$ on the left and $(0^{\circ}, 0^{\circ}, 0^{\circ})$ on the right.
\textbf{(a)} We see clear gradients toward the two basins of attraction. There is less change along the elevation due to less change at each step. \textbf{(b)} The agent can also change the lighting of the scene, with qualitatively similar results. In this graphic we only measure the absolute value of the distance.
}
\label{fig:all_seismic.png}
\end{adjustwidth}
\end{figure}

The heat distribution in this case is more noisy. To get a view of the expected heat map profile, we average several figures of this type to show basins of attraction during the search. Each heat map is shifted such that the goal position is at the bottom, middle row (Fig~\ref{fig:all_seismic.png}, a).
Here it is obvious that the goal and the $180^{\circ}$ flipped (azimuth)  version of the goal are attractor states. This is due to the representation map being sensitive to the rough shape of the object, but being unable to distinguish finer details. In (Fig~\ref{fig:all_seismic.png}, b) we display an aggregate heat map when the agent can also change the lighting conditions. 

Our visualizations show a gradient toward the goal state in addition to visually similar far-away-states,\added{ sometimes causing the algorithm to produce solutions that are the polar opposite of the goal concerning the azimuth}. Prediction errors \added{also} prevent the planning algorithm from finding the exact goal for every task\added{, even if it is not distracted by the polar-opposite.} \deleted{ sometimes producing solutions that are the polar opposite of the goal concerning the azimuth.}

\begin{figure}
\begin{adjustwidth}{-2.25in}{0in} 
\centering
 \includegraphics[width=1.25\textwidth]{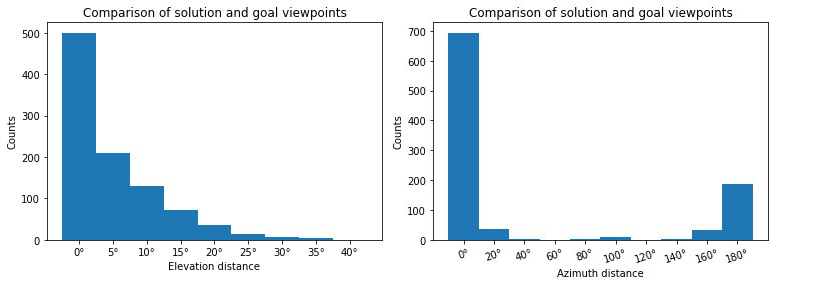}
\caption{ {\bf Histograms of elevation-wise and azimuth-wise VGG16 errors.} The histograms display the counts of the distance between goal and solution states along elevation (left) and azimuth (right) on test data. The distance between the start and goal viewpoints is equally distributed across all the trials, along both dimensions. The goal and the $180^{\circ}$ flipped (azimuth) version of the goal are attractor states.
}
\label{fig:allinone_histos.png} 
\end{adjustwidth}
\end{figure}

To investigate the accuracy of the search with respect to each dimension separately, we plot the histogram of distances between the goal states and the solution states in Fig~\ref{fig:allinone_histos.png}. The goal and start states are chosen randomly, with the restriction that the azimuth distance and elevation distance between them are each uniformly sampled.  For the rest of the paper, all trials follow this sampling procedure.  The results look less accurate for elevation than azimuth because the elevation changes are smaller than the azimuth changes in the NORB data set.  The difference between the goal and solution viewpoints in Fig~\ref{fig:allinone} left, for example, is hardly visible. If one would scale the histograms by angle and not by bins, the drop-off would be similar.

\subsection*{Latent space dimensionality}
With the next experiment, we aim to answer \added{research question 2 (trained predictability). While tuning the details of the design, we also tackle} research questions 3 (dimensionality) and 4 (solution constraints). 
We do an ablation study of the dimensionality of the representation for our method  (Table~\ref{featuresablation}). The test car is an unseen car toy from the NORB data set and the train car comes from the training set. 

\begin{table}\begin{center}
  \captionof{table}{{\bf Ablation study of the representation dimensionality.} We change the output dimension of the representation learner subnetwork and compare it to the VGG16 representation trained on ImageNet. The performance (mean success rate) is averaged over ten separate instantiations of our systems, where each instance is evaluated on a hundred trials of the viewpoint-matching task. A trial is a success if the goal is reached by taking less than twice the minimum number of actions needed to reach it. The standard deviations range between 0.1 and 0.3 for each table entry.}
  \label{featuresablation}
 \begin{tabular}{l r r r r} 
 Representation & Dimensions & Training Car ($\%$) & Test Car ($\%$)  \\ [0.2ex] 
\arrayrulecolor{gray} \hline\\[-2.5ex]LARP (Contrastive)  & 96  &59.3& 56.8 \\ 
     & 64 &  64.1& 60.5\\ 
    & 32&72.3& 59.4  \\ 
     & 16 &74.1& 59.3  \\
  & 8&  82.7& 58.0   \\ [0.1ex]
  \hline\\[-2.5ex]LARP (Sphering) & 96 & 41.1& 37.8  \\ 
 &  64& \textbf{93.9}&53.7  \\ 
  & 32&  89.8 & 51.9  \\ 
 &16&85.2&42.6  \\
       &  8 &85.1    & 40.1 \\ [0.1ex] 
   \hline\\[-2.5ex]LARP (Decoder)     &  96& 58.0    &51.9  \\ 
   &   64& 79.5 & \textbf{63.0}  \\ 
  &  32   & 77.8 & 61.7\\ 
  &  16   &51.9 & 45.2 \\ 
  &  8  &51.1   & 42.4  \\ [.1ex] 
  \hline\\[-2.5ex]VGG16 &  902  & 62.4 & 55.1 \\ [.1ex]
 \hline\\[-2.5ex]Random Steps &  & 3.5  & 3.5 \\ [1ex] 
\end{tabular}
\end{center}\end{table}

There is no complete and total winner: the network with the sphering layer does the best on one of the cars used during training while the reconstructive-loss network does the best on the held-out test car. The sharp difference in performance between 64 and 94 sphering-regularized representation can be explained by the numerical instability of the power iteration method for too large matrix dimensions. 

The VGG16 representation is not the highest performer on any of the car toys. Many of VGG16's representation values are 0 for all images in the NORB data set, so we only use those that are nonzero for any of the images. We suggest that this high number of dead units is due to the representation being too general for the task of manipulating relatively homogenous objects. Another drawback of using pre-trained networks is that information might be encoded that is unimportant for the task. This has the effect that our search method is not guaranteed to output the correct solution in the latent space, as there might be distracting pockets of local minima. 

The random baseline has an average success rate of 3.5$\%$, which is very clearly outperformed by our method.  As 64 is the best dimensionality for the representation on average, we continue with that number for our method in the transfer learning experiment.

\added{We conclude that the proposed method of training a representation for predictability is feasible.} So far we have evidence that 64 is the best dimensionality of the \added{representation's} latent space for our planning tasks. However, it is not yet conclusive what the best restriction is to place on the representation to avoid the trivial solution, in terms of planning performance. 

\subsection*{Comparison with other RL methods}
Now we divert our attention to research question 5 (benchmarking), where we compare our method to the literature. 
 For the comparison with standard RL methods, we use the default configurations of the model-free methods DQN (off-policy) and PPO (on-policy) as defined in OpenAI Baselines~\cite{baselines}. Our model-based comparison is chosen to be world models \cite{ha2018world} from \url{https://github.com/zacwellmer/WorldModels}. We make sure that the compared RL methods are similar to our system in terms of the number of parameters as well as architecture layout and compare them with our method on the car viewpoint-matching task.

 
 \added{The task setup is the same as before and is converted to an OpenAI gym environment: a start observation and a goal observation are passed to the agent. If the agent manages to reach it within 2 times the minimum number of actions required (the minimum number is calculated by the environment), the agent receives a reward and the task is considered a success. Otherwise, no reward is given. }

\begin{figure}
\begin{adjustwidth}{-1.91in}{0in}
    \centering
    
    \subfloat[{\bf DQN Performance.}]{{\includegraphics[width=6.3cm, angle=90]{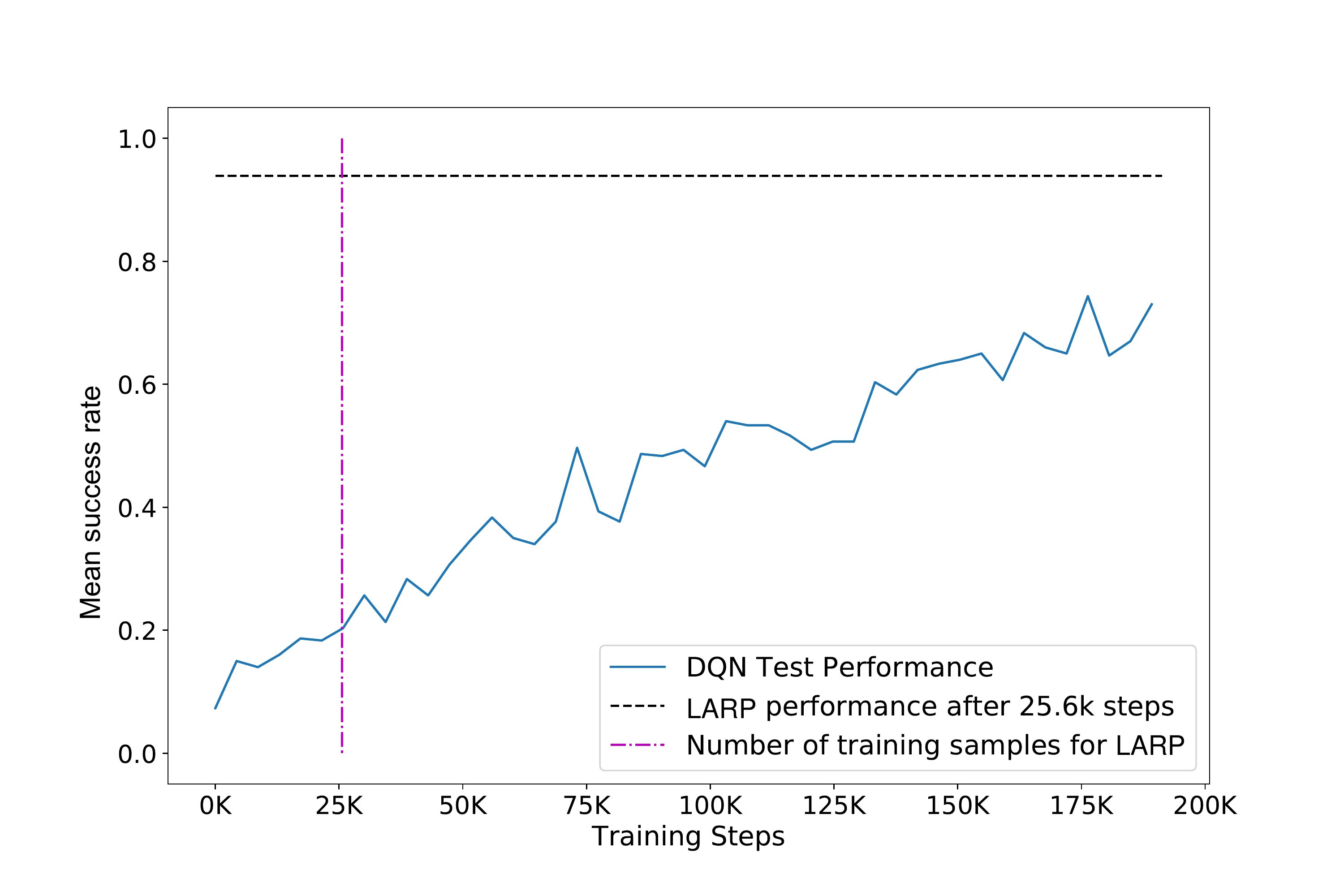}}}
    \qquad
      \hspace{-5em}
    \subfloat[{\bf PPO Performance.}]{{\includegraphics[width=6.3cm, angle=90]{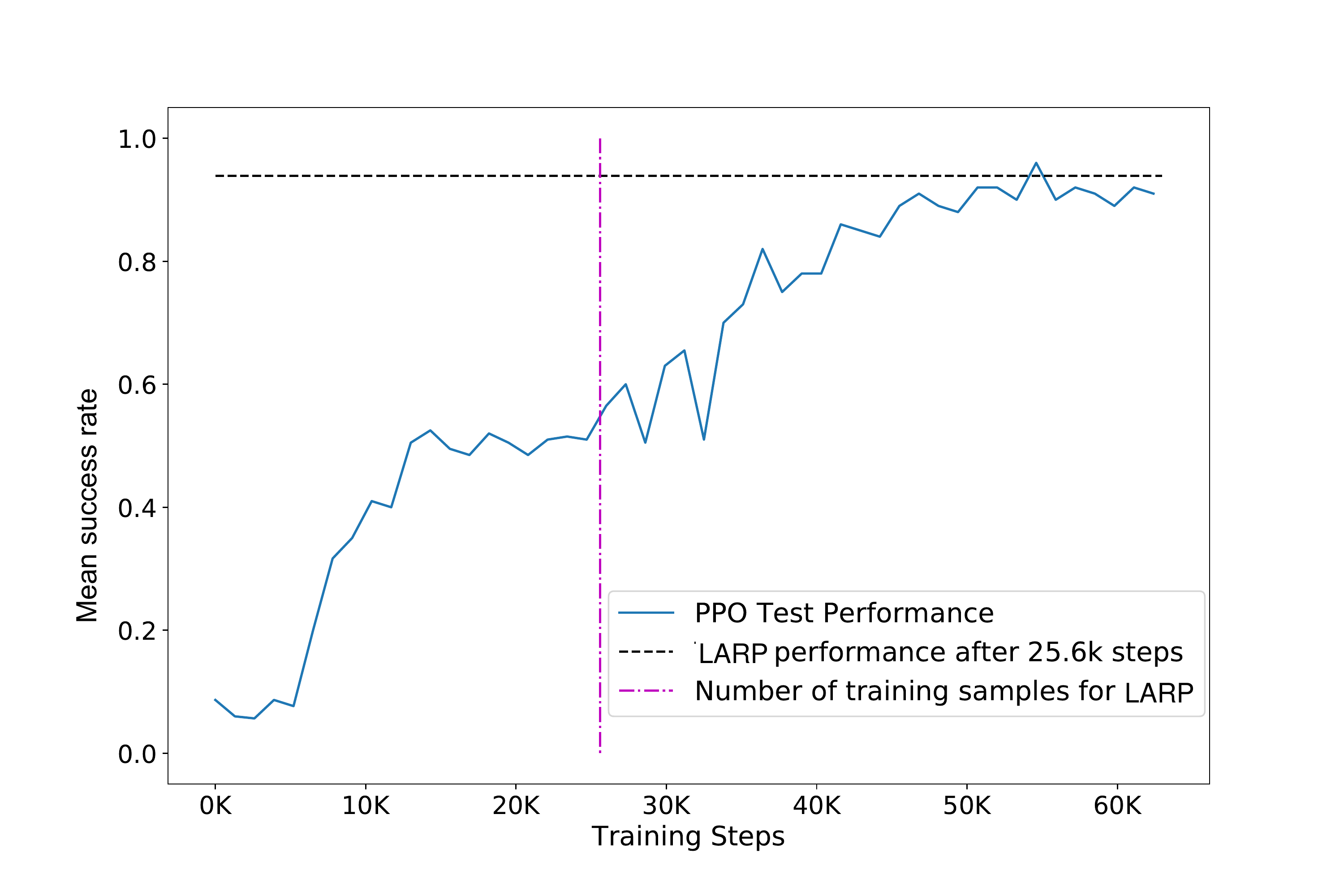}}}
    \qquad
      \hspace{-5em}
      
    \subfloat[{\bf World Models Performance.}]{{\includegraphics[width=6.3cm, angle=90]{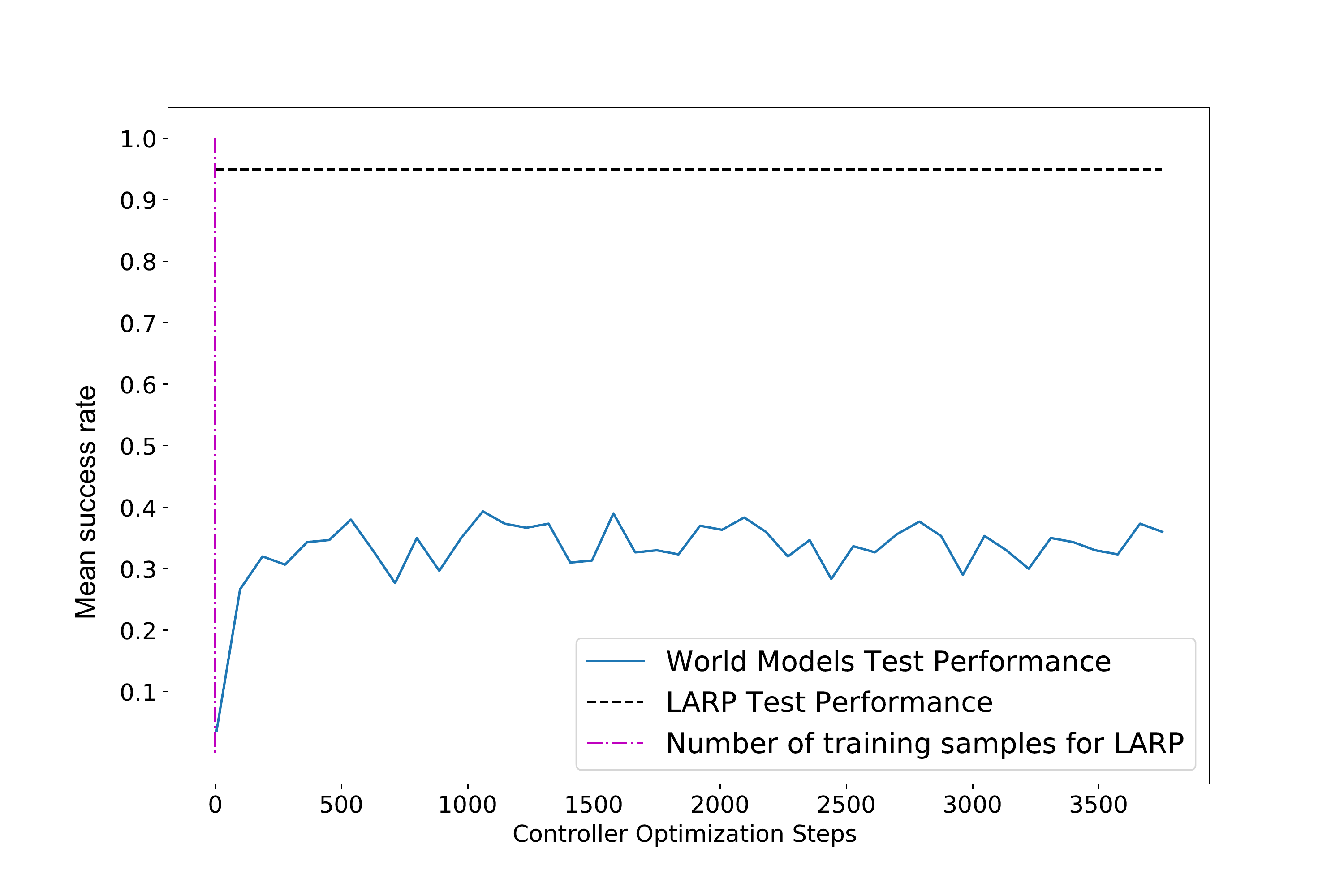}}}
    \qquad
      \hspace{-5em}
    \qquad
      \hspace{-5em}
    \caption{{\bf Reinforcement learning comparison.} The vertical dashed lines indicate when the compared algorithm has processed the same number of transitions as our method and the horizontal dotted line indicates the test performance of our method. Each data point is the   mean \added{success rate} of 100 test episodes \added{after a varying number of training steps}, averaged over 5 different seeds of each learner. The model-free methods in {\bf(a)} and {\bf(b)} train the representation and the controller simultaneously by acting in the environment and collecting new experiences. The representation in {\bf(c)} is trained on 25.6k transitions, which is the same number we use. The plot shows the optimization curve for the controller, using a Covariance-Matrix Adaptation Evolution Strategy, which hardly improves after 500 or so training steps. The horizontal line starts at 0 for world models because the representation has finished training on the observations before the controller is optimized. }
    \label{fig:rlfigs}
\end{adjustwidth}
\end{figure}

 The results of the comparison can be viewed in Fig.~\ref{fig:rlfigs}. \added{Each point in the curve contains each method's mean success rate: the average of the cumulative reward from 100 test episodes from 5 different instantiations of the RL learner, so it is the average reward over 500 episodes in total. The test episodes are done on the same environment as is used for training, except that the policy is maximally exploiting and minimally exploring.}

 \replaced{In our experiments, the}{The} DQN networks are much more sample inefficient than PPO, which in turn is more sample inefficient than our method. However, our method is more time-consuming during test time. We require a forward pass of the predictor network for each node that is searched before we take the next step, \added{which can grow rapidly if the target is far away}. \replaced{In contrast, only a single pass through the traditional RL networks is required to compute the next action.}{ This is  contrasted with a single pass in total of the DQN / PPO networks for the next step.}

Our method reaches $93.9\%$ success rate on the train car (Table~\ref{featuresablation}) using 25.6k samples, but the best PPO run only reaches $70.5\%$ after training on the same number of samples. The best single PPO run needed 41.3k samples to get higher than $93.9\%$ success rate and the average performance is higher than $93.9\%$ at around 55k samples. After that, some PPO learners declined again in performance. The world models policy quickly reaches the same level of performance as DQN got after 50k steps and PPO after approximately 8k steps, but it doesn't improve beyond that.

\begin{figure}
\begin{adjustwidth}{-1.39in}{0in}
    \centering
    
    \subfloat[ {\bf}  ]
    {{\includegraphics[width=6.55cm]{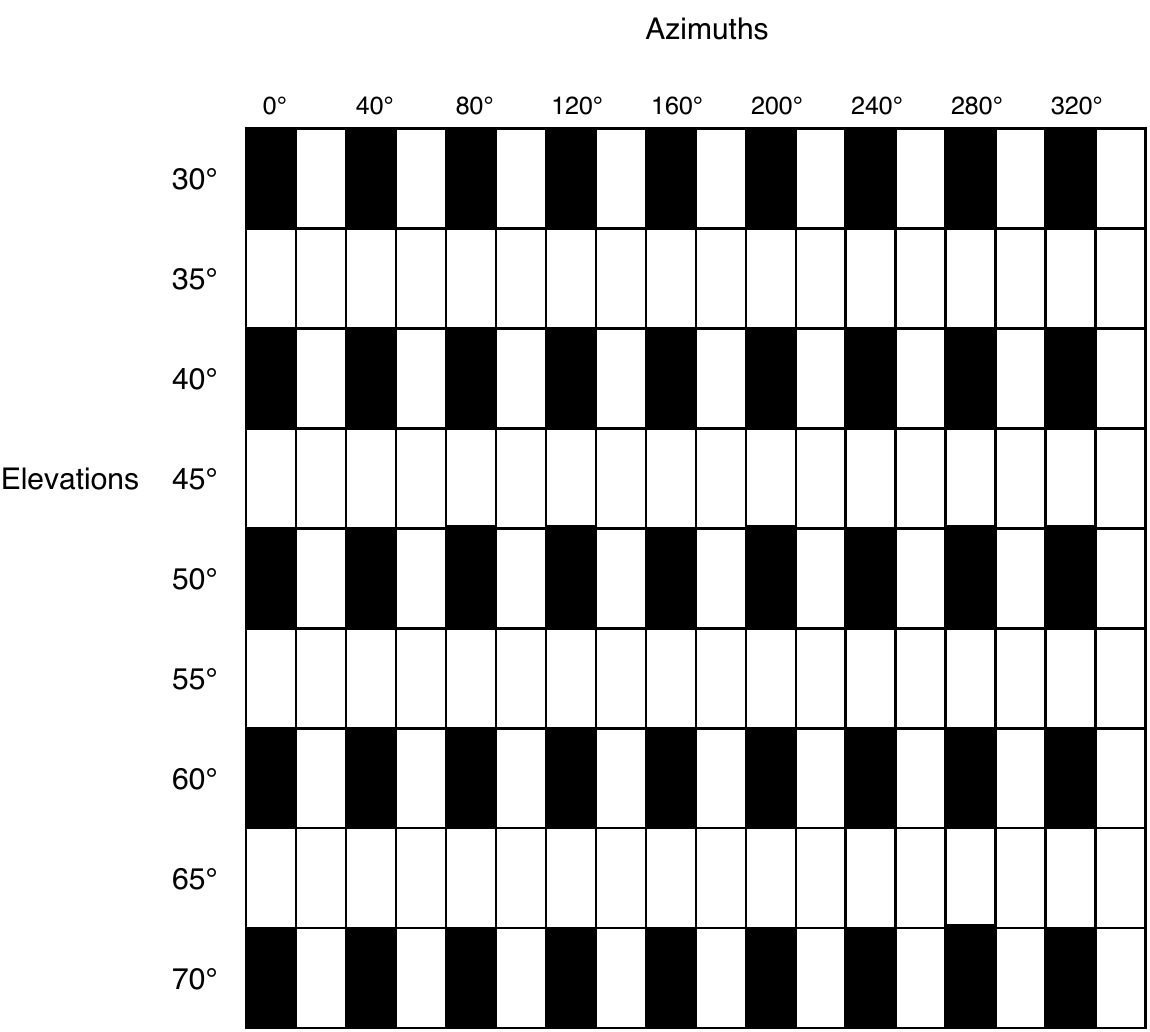}}}
    \qquad
      \hspace{-1.5em}
    \subfloat[{\bf}   ]
    {{\includegraphics[width=9.63cm]{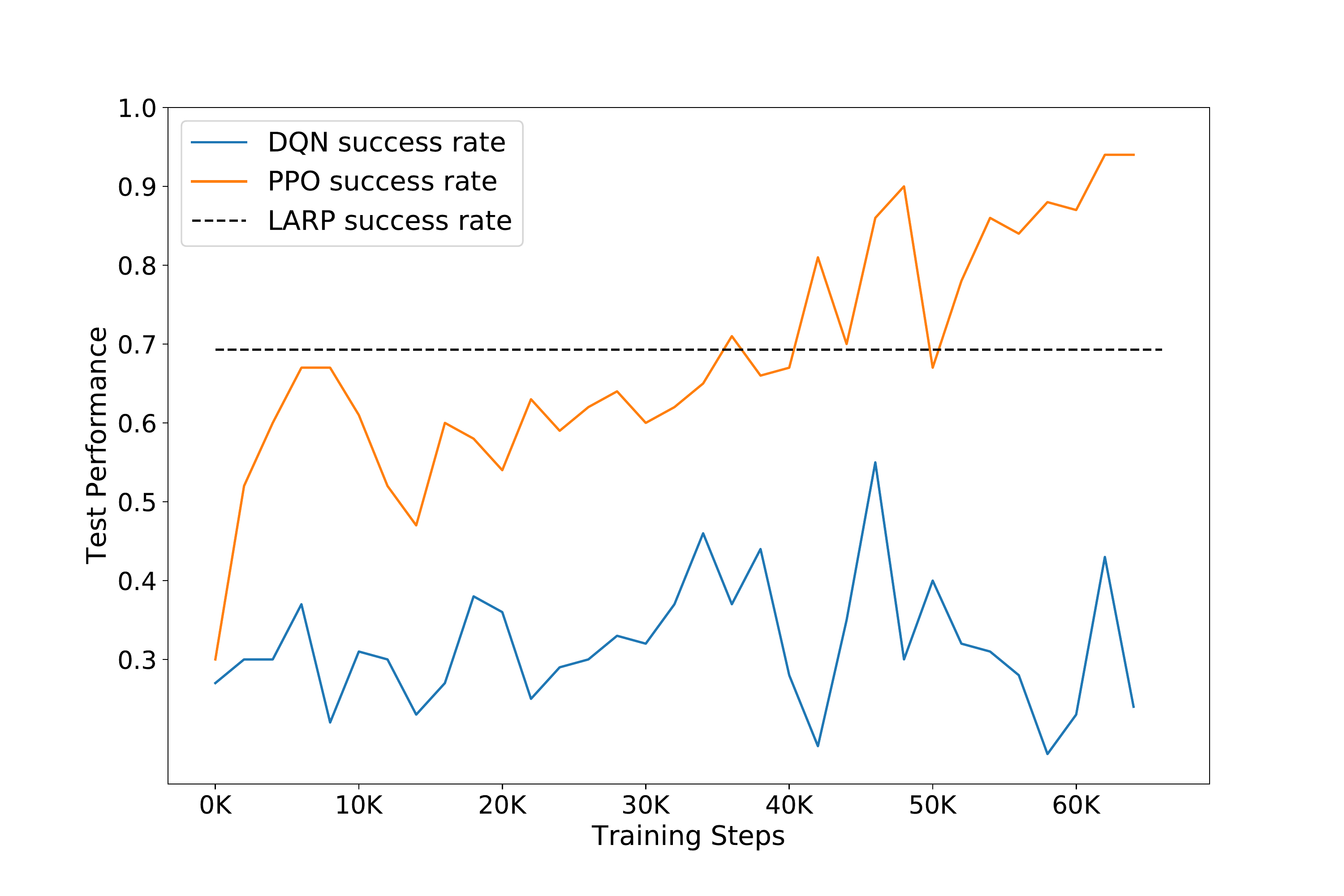}    \vspace{-2.0em}}}
    \caption{{\bf Re-training after placing checkerboard-patterned obstacles.} {\bf (a)} The task is the same as before, but nothing happens if the agent attempts to move to a state containing a black rectangle. {\bf (b)}  After training the agents, we re-tested them after we  introduced the checkerboard pattern of obstacles. Our method does not allow for re-training in the new environment.}
    \label{fig:checkz}
\end{adjustwidth}
\end{figure}

We conclude that our method compares favorably to other methods from the RL literature in terms of sample efficiency.

\subsection*{Modifying the environment}
We now modify the environment to answer research question 6 (generalization).  To see how the methods compare when obstacles are introduced to the environment, we repeat the trial on one of the car objects except that the agent can no longer pass through states whose elevation values are divisible by 10 and azimuth values are divisible by 40 (Fig.~\ref{fig:checkz}, (a)).


As before, the goal and start locations can have any azimuth-elevation pair but the agent cannot move into states with the properties indicated by the black rectangle. Every action is available to the agent at all locations as before, but the agent's state is unchanged if it attempts to move to a state with a black rectangle. 

We trained LARP using the contrastive loss, PPO, and DQN agents until they reached $80\%$ accuracy on our planning task and then tested them with the added obstacles. 
Our method loses about $10\%$ performance, but PPO loses $50\%$. Nevertheless, we can continue training PPO until it quickly reaches top performance again (Fig.~\ref{fig:checkz}, (b)). Our method is not re-trained for the new task and DQN did not reach a good performance again in the time we allotted for re-training.  

Thus, we see that our method is quite flexible and generalizes well when obstacles are introduced to the environment.

\subsection*{Transfer to dissimilar objects}

\replaced{We now consider research question 6 (generalization) further by investigating how well our method transfers knowledge from one domain to another. 
}{We now investigate how well our method transfers knowledge from one domain to another, seeking to answer research question 7.}
 
\begin{figure} 
\begin{adjustwidth}{-0.2in}{0in}
\centering
 \captionsetup{width=.8\linewidth}
  \resizebox*{1.   \textwidth}{!}{\includegraphics
{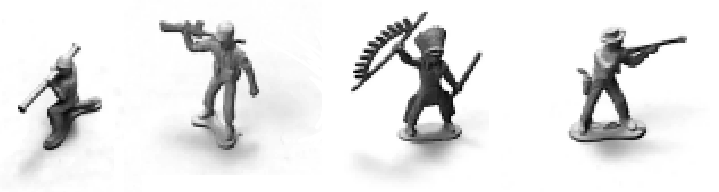}}
\caption{ {\bf Toys for transfer learning experiments.} From left to right: Soldier (Kneeling), Soldier (Standing), Native American with Bow and Cowboy with Rifle. 
}
\label{fig:merge4.png}
\end{adjustwidth}
\end{figure}

Selecting the best dimensionality for the representation from the previous set of experiments, we investigate further their performance in harder situations using unseen, non-car objects. \replaced{The models are trained on the same car objects as in the previous experiment, but they are tested on an array of different plastic soldiers: a kneeling soldier holding a bazooka, a standing soldier with a rifle, a native american with a bow and spear and a cowboy with a rifle (Fig.~\ref{fig:merge4.png}).}{(Table~\ref{moretesttoys}).}

\subsubsection*{\added{Qualitative results}}

\begin{figure}[ht!]
\begin{adjustwidth}{-0.93in}{0in}
\captionsetup[subfigure]{justification=centering,singlelinecheck=false}
\begin{subfigure}{0.41\textwidth}
\includegraphics[width=\linewidth]{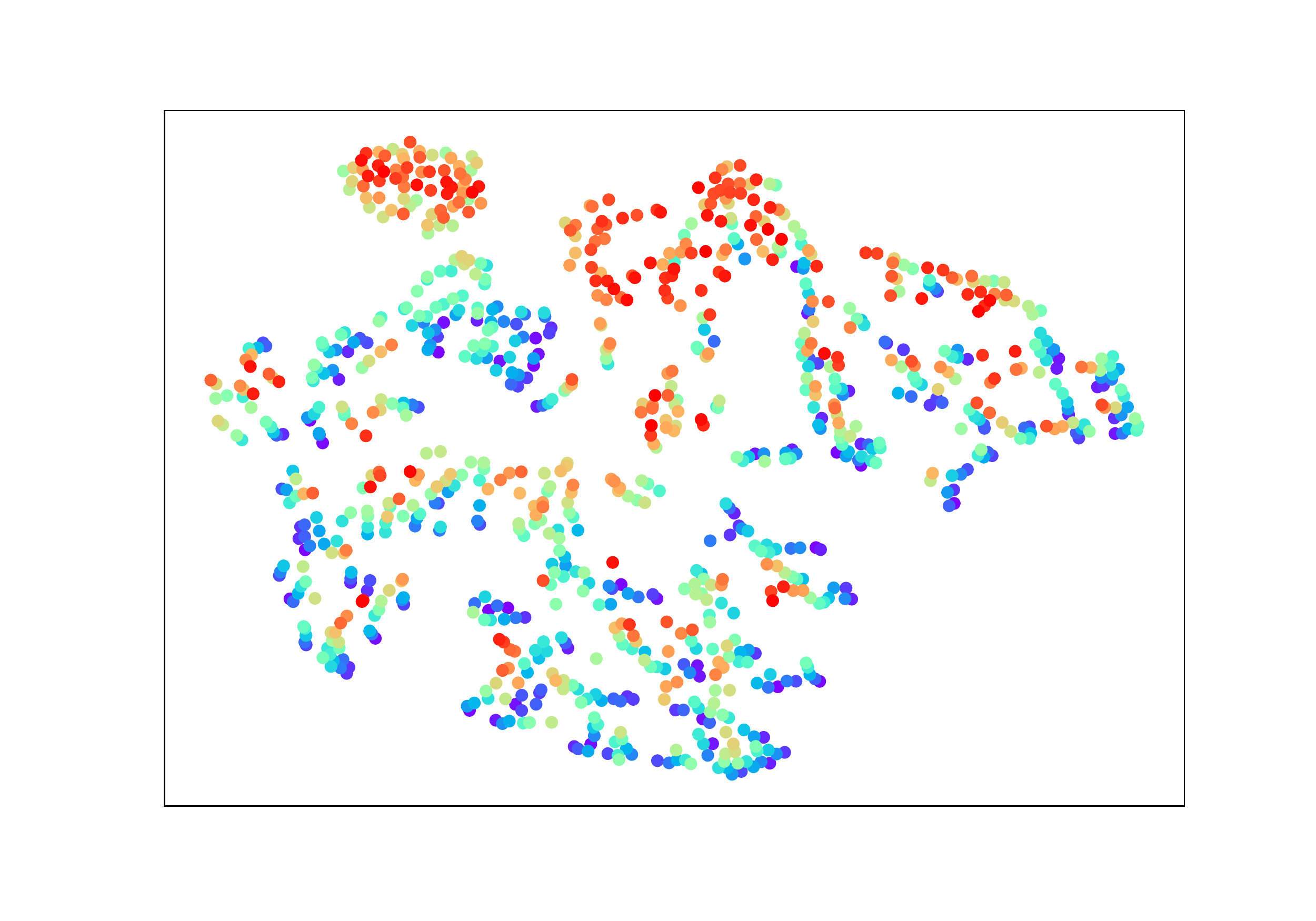} 
\caption{\added{LARP Elevation t-SNE}} \label{fig:1picsa}
\end{subfigure} 
\hspace{-2em}
\begin{subfigure}{0.41\textwidth}
\includegraphics[width=\linewidth]{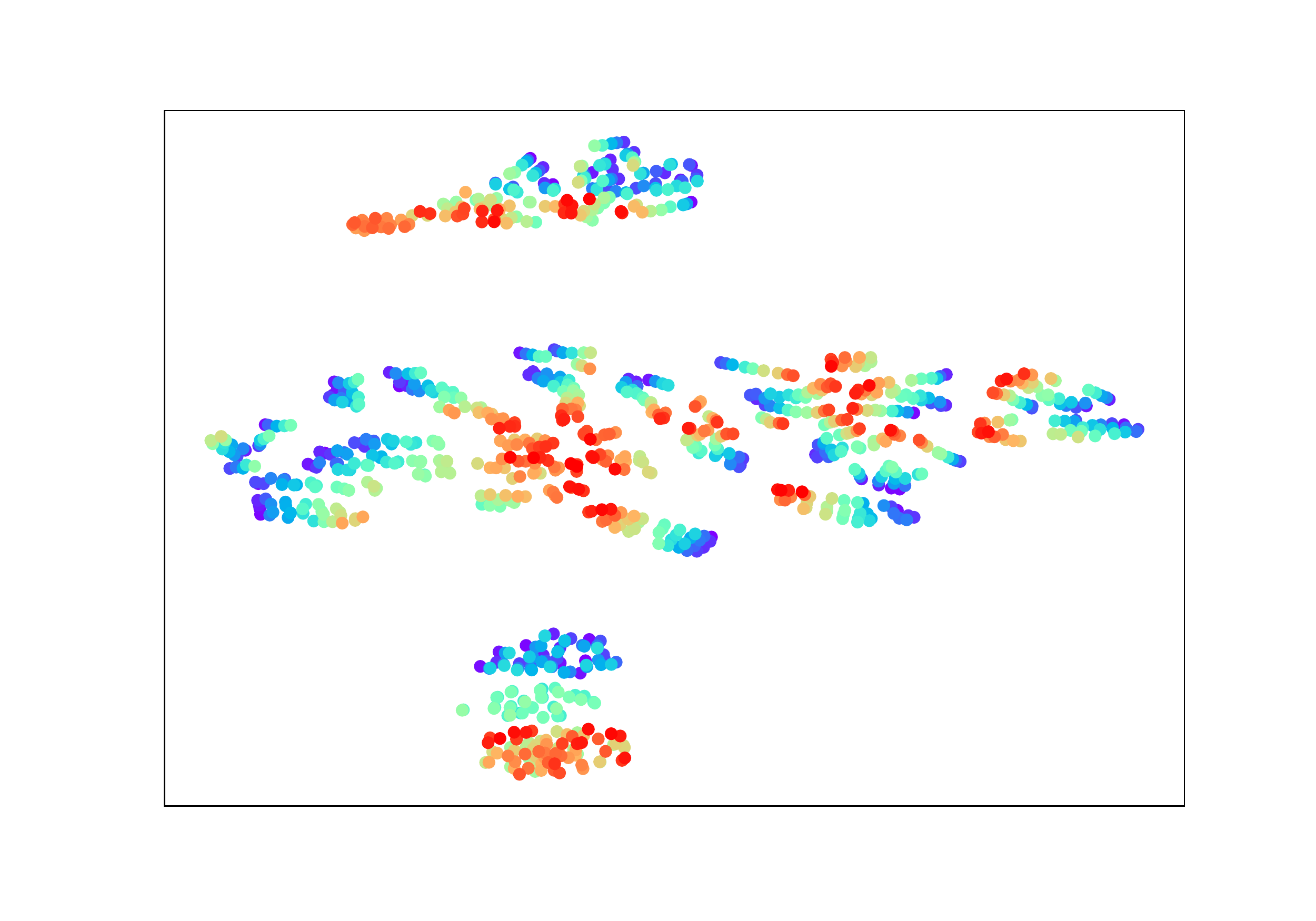}
\caption{\added{Conv. Encoder Elevation t-SNE}} \label{fig:2picsb}
\end{subfigure}
\hspace{-2em}
\begin{subfigure}{0.41\textwidth}
\includegraphics[width=\linewidth]{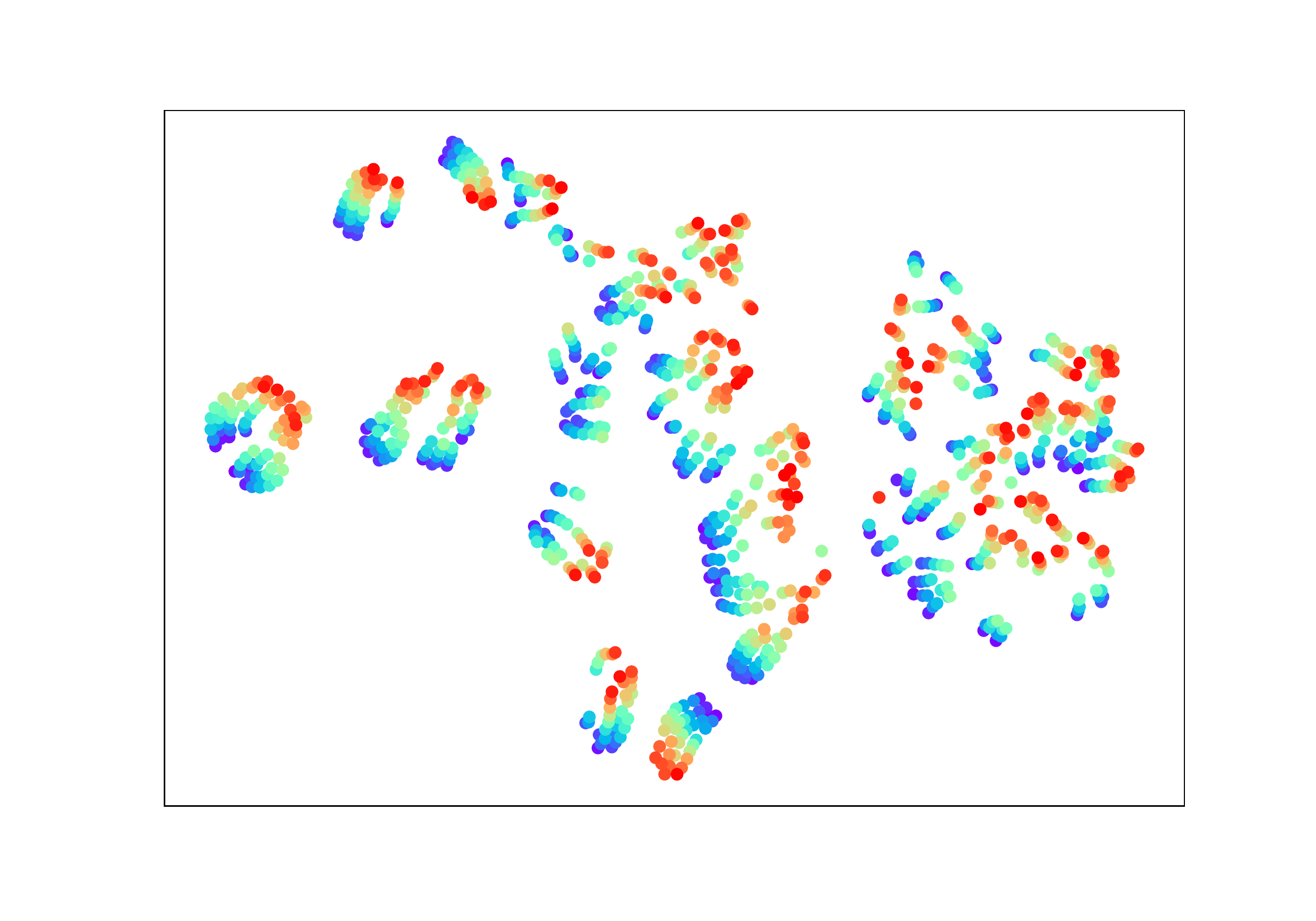}
\caption{\added{VGG16 Elevation t-SNE}} \label{fig:3picsb}
\end{subfigure}
\smallskip
\begin{subfigure}{0.41\textwidth}
\includegraphics[width=\linewidth]{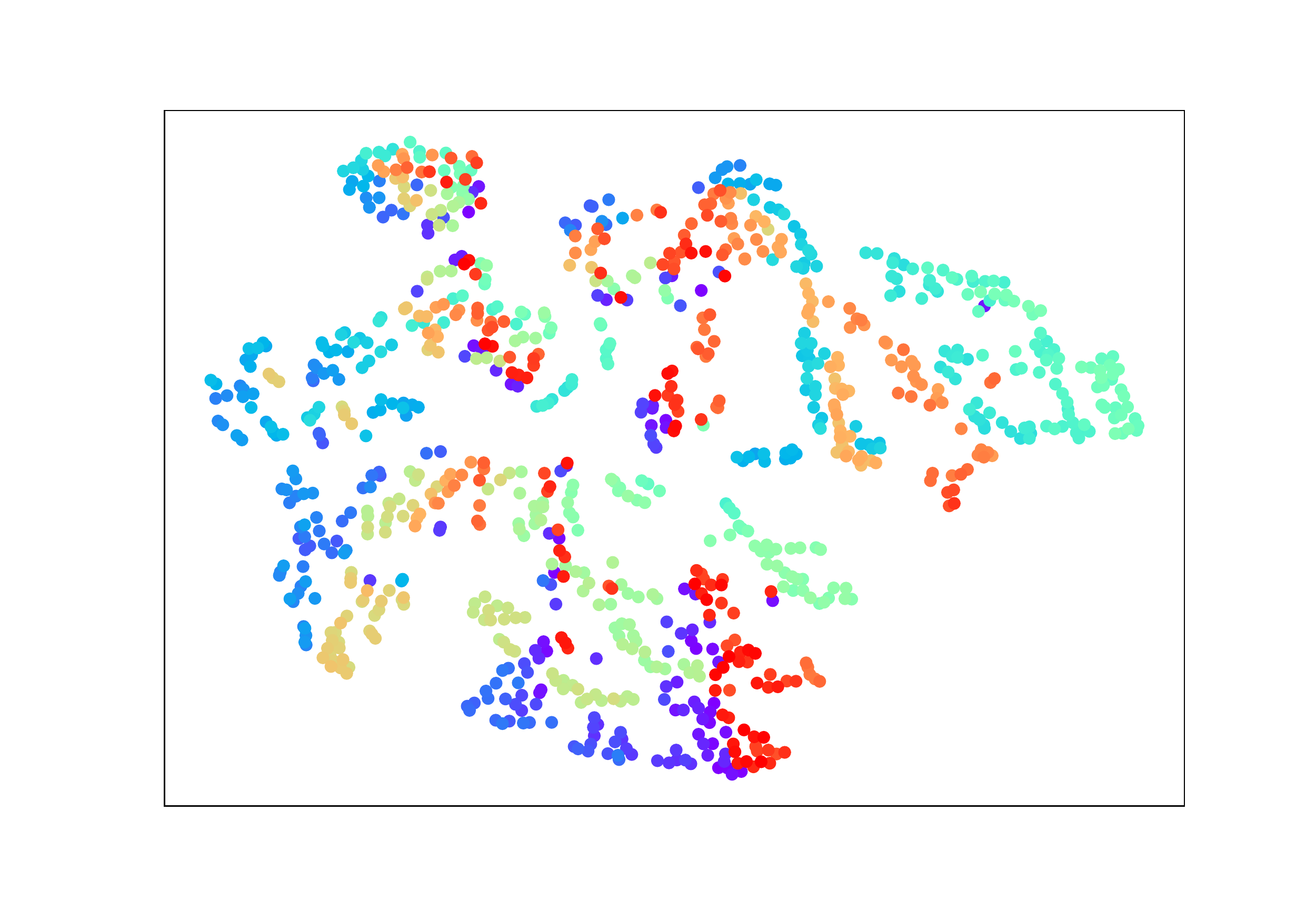} 
\caption{\added{LARP Azimuth t-SNE}} \label{fig:4picsa}
\end{subfigure} 
\hspace{-2em}
\begin{subfigure}{0.41\textwidth}
\includegraphics[width=\linewidth]{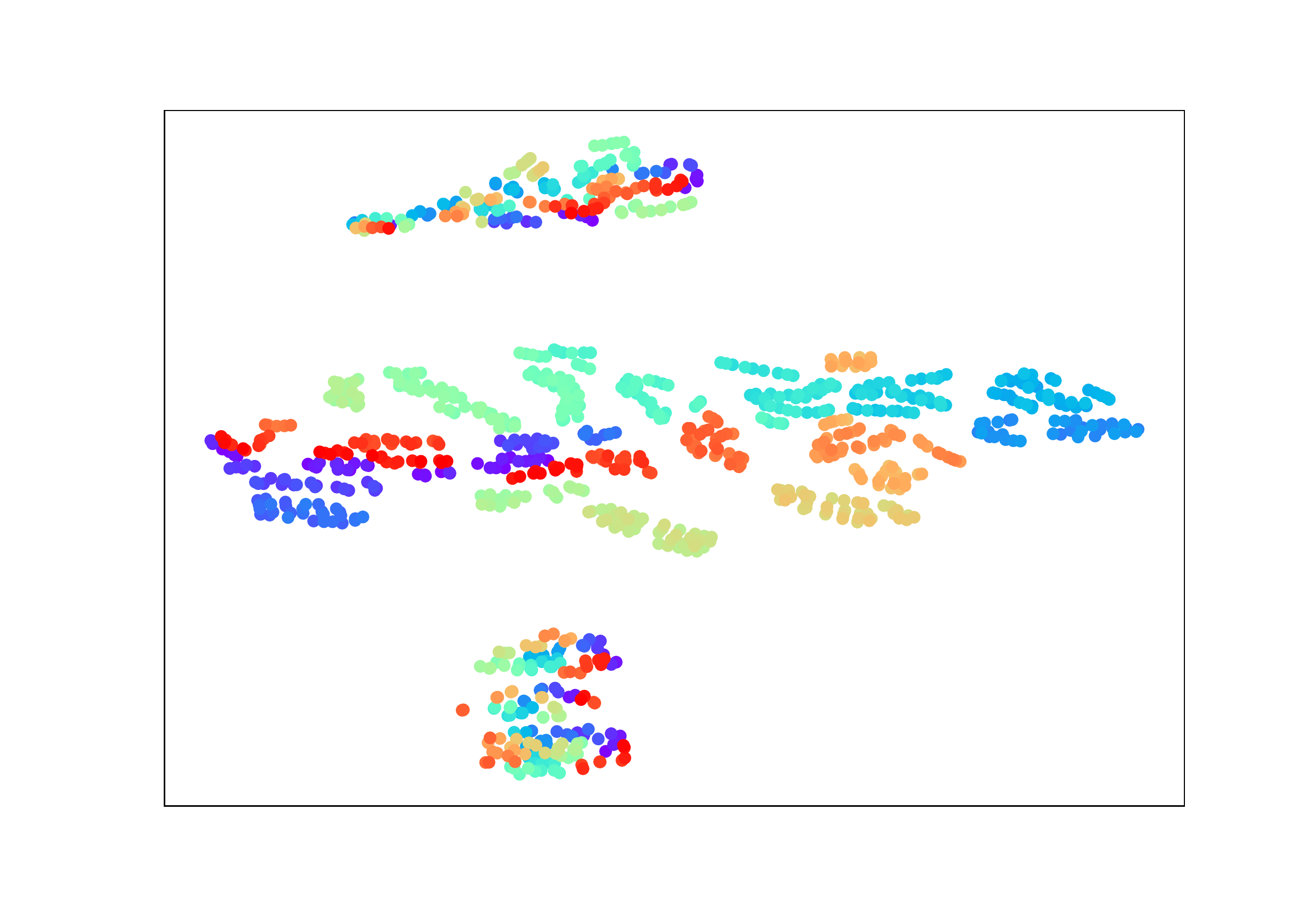}
\caption{\added{Conv. Encoder Azimuth t-SNE}} \label{fig:5picsb}
\end{subfigure}
\hspace{-2em}
\begin{subfigure}{0.41\textwidth}
\includegraphics[width=\linewidth]{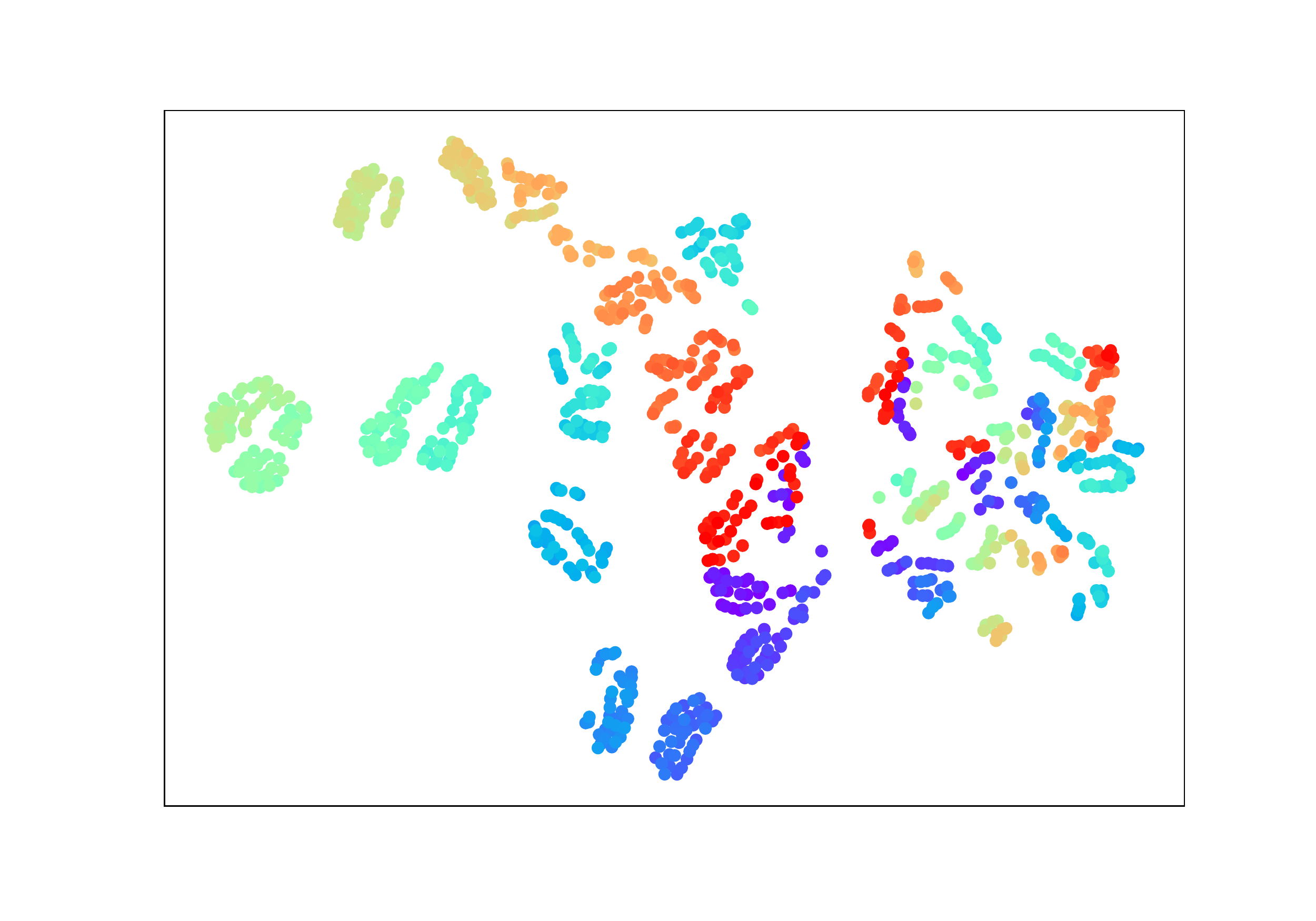}
\caption{\added{VGG16 Azimuth t-SNE}} \label{fig:6picsb}
\end{subfigure}
\smallskip
\begin{subfigure}{0.41\textwidth}
\includegraphics[width=\linewidth]{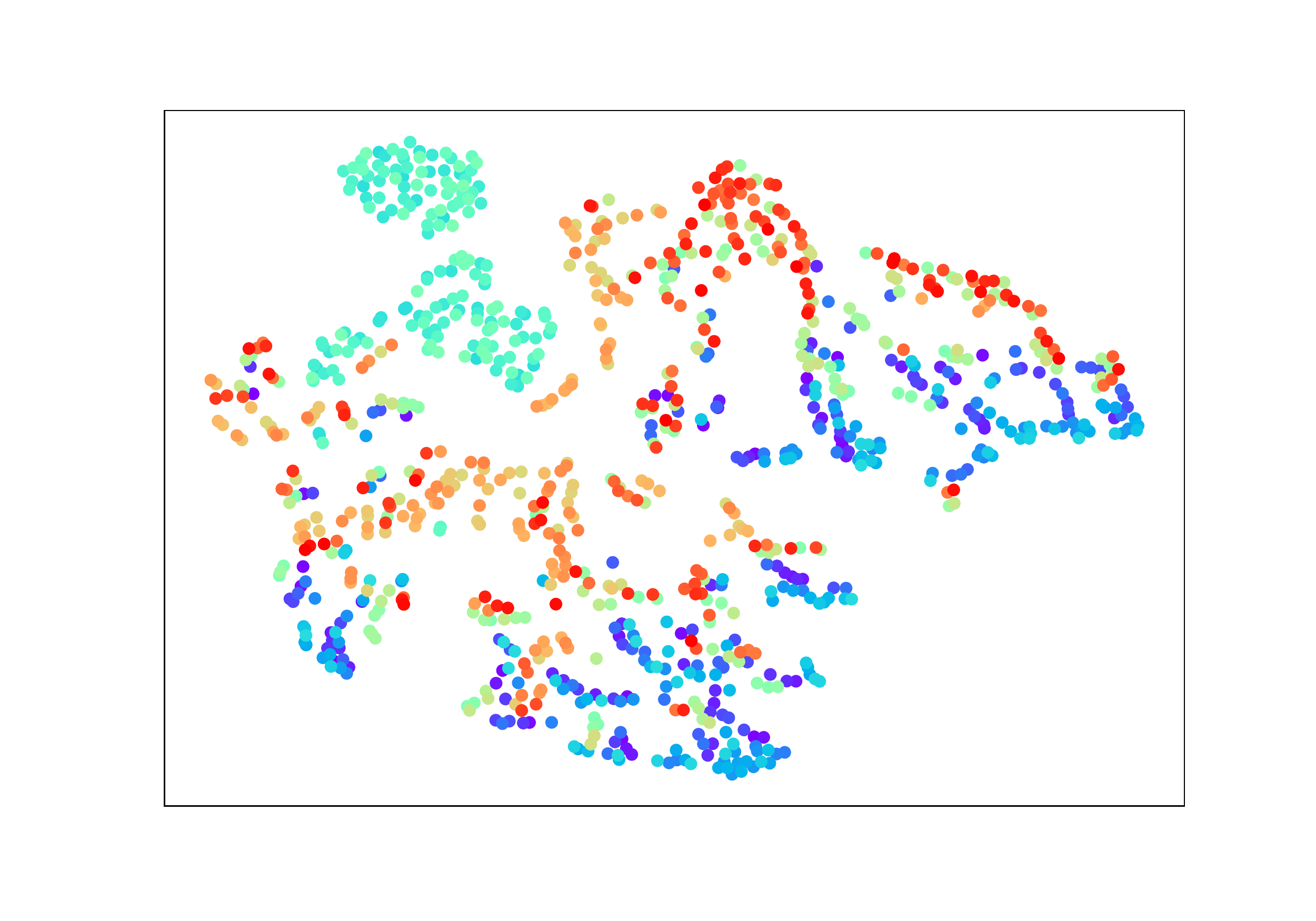}
\caption{\added{LARP Lighting t-SNE}} \label{fig:7picsc}
\end{subfigure}
\hspace{-2em}
\begin{subfigure}{0.41\textwidth}
\includegraphics[width=\linewidth]{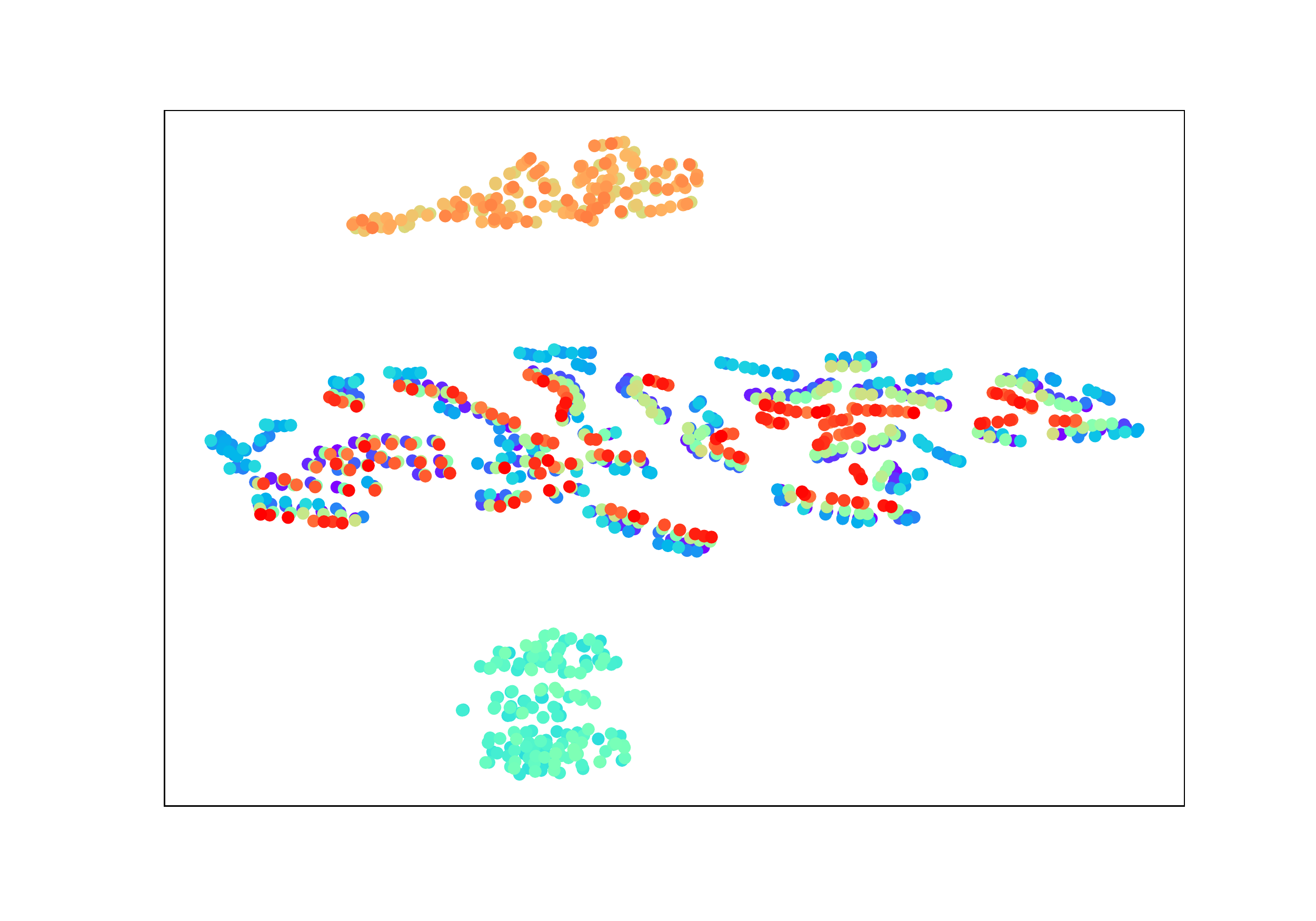}
\caption{\added{Conv. Encoder Lighting t-SNE}} \label{fig:8picsd}
\end{subfigure}
\hspace{-2em}
\begin{subfigure}{0.41\textwidth}
\includegraphics[width=\linewidth]{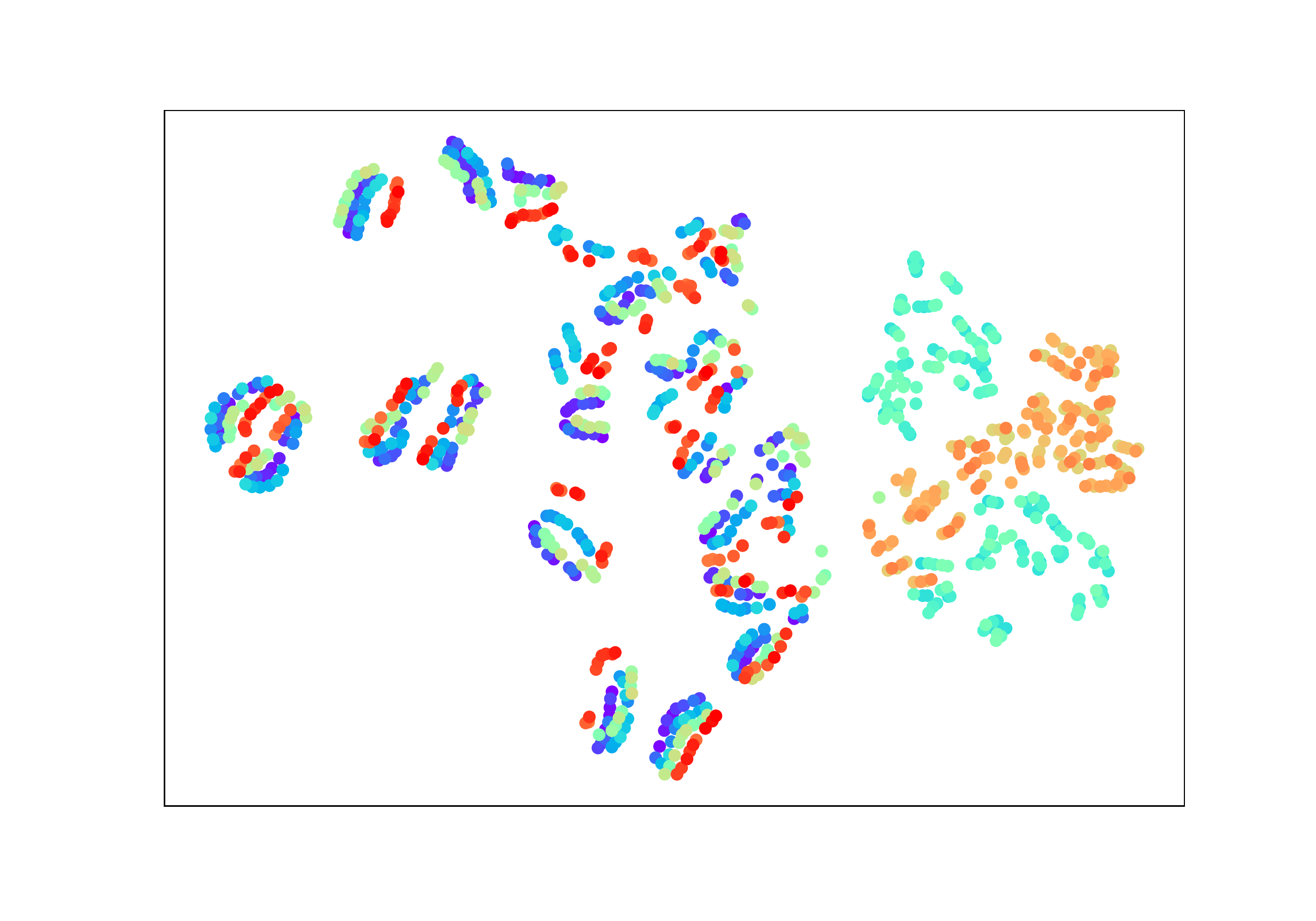}
\caption{\added{VGG16 Lighting t-SNE}} 
\label{fig:9picsb}
\end{subfigure}
\captionsetup{margin=0cm}
\caption{ \added{The clustering structure of a toy soldier embedding after dimensionality reduction with t-SNE. The top row colors the samples by elevation, the middle row by azimuth and the bottom one by lighting. The left column is the embedding given by our method using the contrastive loss term, the middle by a convolutional encoder, and the right by a pre-trained VGG16 network.}} \label{fig:4apics}
\end{adjustwidth}
\end{figure}

\added{We offer a visualization of the learned representation of the kneeling soldier toy using our method, a convolutional encoder, and VGG16  in Fig.~\ref{fig:4apics}. Each embedding was reduced to 2 dimensions using t-SNE. }

Every method structures the domain in similar way. In the bottom row, we see that largest clusters for all methods are the ones with the highest (teal dots) illumination settings, which is explained by the effect of the lighting on the pixel value intensities. Within these clusters, we see clustering based on the azimuth (middle row). Finally, within these clusters, there is a gradient structure based on elevation (top row). This is due to the elevation changing in smaller step-sizes, with 5 degree differences, than azimuth with 20 degree differences. 

\subsubsection*{\added{Quantitative results}}

Our experiments include the pre-trained VGG16 network because we believe that a flexible algorithm should rather be based on a generic multi-purpose-representation and not on a specific representation.  To our surprise, it is outperformed by our approach on every object, even as the viewpoint-matching task is done on objects that are different from the original car toys they are trained on.

We also try transfer learning \deleted{a} VAE \added{and a convolutional encoder} using the same network as our representation \replaced{and a UMAP embedding, with 64 features each.}{ with 64 features each and a UMAP embedding.} \added{We display the results of the same viewpoint-matching tasks as before using these different representations on the unseen army figures in Table~\ref{moretesttoys}.} \replaced{Convolutional encoders have}{UMAP has}  the lowest performance out of all the methods but the VAE representation \replaced{has the most similar performance to}{rivals} our own representations.

\begin{table}\begin{center}
\captionof{table}{ {\bf Transfer learning performance.} The methods are trained on the data set of different car images and then their performance on dissimilar toys is tested. Each number is the mean success rate of viewpoint matching out of 1000 trials.} 
\label{moretesttoys}
 \begin{tabular}{ l r r r r | r   l} 
  Representation & Soldier & Soldier & Native American   & Cowboy  & 
 \\ [0.5ex] 
 &   (Kneeling)  &  (Standing)  & with Bow & with Rifle & \added{Mean}  \\ [0.5ex] 
 \arrayrulecolor{gray}\hline 

   LARP (Contrastive)    &  \textbf{\fpeval{round(100*(23+18+18+21+12)/(5*81), 1)}} & \fpeval{round(100*(11+8+12+5)/(4*81), 1)} & \fpeval{round(100*(11+11+10+7+0.11)/(4*81), 1)} & \textbf{\fpeval{round(100*(17+17+15+18)/(4*81), 1)}} & \added{\textbf{16.7}}\\
    LARP (Sphering)   &  \fpeval{round(100*(14.9)/(1*81), 1)} &  \fpeval{round(100*(12.43)/(1*81), 1)}&  \textbf{\fpeval{round(100*(12.34)/(1*81), 1)}} & \fpeval{round(100*(12.8)/(1*81), 1)}& \added{16.2}\\
    LARP (Decoder) &  \fpeval{round(100*(11.6)/(1*81), 1)} &  \textbf{\fpeval{round(100*(13.43)/(1*81), 1)}} &  \fpeval{round(100*(11.93)/(1*81), 1)} & \fpeval{round(100*(12.63)/(1*81), 1)}& \added{15.3}\\ 
    VAE   &  \fpeval{round(100*(11.7)/(1*81), 1)} &  \fpeval{round(100*(10.61)/(1*81), 2)} &  \fpeval{round(100*(11.01)/(1*81), 1)} & \fpeval{round(100*(11.211)/(1*81), 1)}& \added{13.7}\\
    VGG16   &  \fpeval{round(100*(10)/(1*81), 1)} &  \fpeval{round(100*(10.45)/(1*81), 1)} &  \fpeval{round(100*(9.54)/(1*81), 1)} & \fpeval{round(100*(11)/(1*81), 1)}& \added{12.7}\\ 
UMAP   &  \fpeval{round(100*(7)/(1*81), 1)} &  \fpeval{round(100*(8.75)/(1*81), 1)} &  \fpeval{round(100*(7.84)/(1*81), 1)} & \fpeval{round(100*(9)/(1*81), 1)}& \added{10.1}\\ 
 
\added{Conv. Encoder}   &  \added{\fpeval{round(100*(4)/(1*81), 1)}} &  \added{\fpeval{round(100*(11.6)/(1*81), 1)}} & \added{\fpeval{round(100*(9.3)/(1*81), 1)}} & \added{\fpeval{round(100*(5.2)/(1*81), 1)}}& \added{9.3} \\ 
    
\end{tabular}\end{center}\end{table}

Our representation is better than the others for this experiment. But we see that the performance of our method drops significantly as we attempt generalizing to different objects with changed input statistics.

\added{Note that even though the representation clustering (Fig.~\ref{fig:4apics}) was more clear-cut using convolutional encoders and VGG, compared to ours, they did not perform as well in the trials. We attribute this to the fact that the main loss term in our representation was the one of future state  predictability, which may not give as clear of a structure in two dimensions as a representation that is trained only for static image reconstruction.}

\section*{Discussion}

In this work, we present Latent Representation Prediction (LARP) networks with applications to visual planning. We jointly learn a model to predict transitions in Markov decision processes with a representation trained to be maximally predictable\replaced{. This allows}{, allowing} us to accurately search the latent space defined by the representation \added{using a heuristic graph traversal / best-first search algorithm}. We validate our method on a viewpoint-matching task derived from the NORB data set. Not surprisingly, a representation that is optimized simultaneously with the predictor network performs best in our experiments.

A common issue of unsupervised representation learning is one of trivial solutions: a constant representation which optimally solves the unsupervised optimization problem but brings across no information. To avoid the trivial solution we constrain the training by introducing a sphering layer or a loss term that is either contrastive or reconstructive. Any of these approaches will do the job of preventing the representations from collapsing to constants, and none of them displays absolute superiority over the others.

Our LARP representation is competitive with pre-trained representations for planning and compares favorably to other reinforcement learning (RL) methods. Our approach is a sound solution for learning a meaningful representation that is suitable for planning only from interactions. 

Furthermore, we find that our method has better sample-efficiency during training than several reinforcement learning methods from the literature\replaced{.}{and our learned representation and predictive model transfer to harder scenarios.}
However, a disadvantage of our method compared to standard RL methods is that the execution time of our method is longer, as a forward pass is calculated for each node during the latent space search\added{, potentially resulting in a combinatorial explosion}.

Our approach is adaptable to changes in the tasks. For example, our search would \replaced{only be slightly}{not be} hindered if some obstacles were placed in the environment or some states were forbidden to traverse through. Furthermore, \replaced{our method is independent of specific rewards or discount rates,}{the reward or discount rates can be easily changed for our method} while standard RL methods are usually restricted in their optimization problems. Often, there is a choice between optimizing discounted or undiscounted expected returns. Simulation/rollout-based planning methods are not restricted in that sense: If reward trajectories can be predicted, one can optimize arbitrary functions of these and regularize behavior. For example, a risk-averse portfolio manager can prioritize smooth reward trajectories over volatile ones.

Future lines of work should investigate further the effect of the different constraints on the end-to-end learning of representations suited for a predictive forward model, as well as considering novel ones. The search algorithm can be improved and made faster, especially for higher-dimensional action spaces or continuous ones. Our network could also in principle be used to train an RL system, for instance, by encouraging it to produce similar outptus as ours and thereby combining sample efficiency with fast performance during inference time.

\section*{Supporting information}

\paragraph*{S1 Supplement}
\label{S1_supplement}
{\bf Environment Code} (ZIP)

\paragraph*{S2 Appendix}
\label{S2_appendix}
{\bf Network Keras Code} (PDF)

\paragraph*{S3 Figure}
\label{S3_supplement}
{\bf Train Cars Sample} (PNG)

\paragraph*{S4 Figure}
\label{S4_supplement}
{\bf Test Car Sample} (PNG)

\section*{Acknowledgments} We would like to thank Zahra Fayyaz, Simon Hakenes, and Daniel Vonk for contributing to discussions related to this project. 

\section*{Author contributions}
\textbf{Conceptualization:} Hlynur Davíð Hlynsson, Tobias Glasmachers, Laurenz Wiskott
\newline \newline
\textbf{Data Curation:} Hlynur Davíð Hlynsson, Merlin Schüler, Robin Schiewer
\newline \newline
\textbf{Formal Analysis:} Hlynur Davíð Hlynsson, Merlin Schüler
\newline \newline
\textbf{Funding Acquisition:} Laurenz Wiskott
\newline \newline
\textbf{Investigation:} Hlynur Davíð Hlynsson
\newline \newline
\textbf{Methodology:} Hlynur Davíð Hlynsson, Laurenz Wiskott
\newline \newline
\textbf{Project Administration:} Hlynur Davíð Hlynsson
\newline \newline
\textbf{Resources:} Laurenz Wiskott, Merlin Schüler
\newline \newline
\textbf{Software:} Hlynur Davíð Hlynsson, Merlin Schüler, Robin Schiewer
\newline \newline
\textbf{Supervision:} Laurenz Wiskott
\newline \newline
\textbf{Validation:} Hlynur Davíð Hlynsson
\newline \newline
\textbf{Visualization:} Hlynur Davíð Hlynsson, Robin Schiewer
\newline \newline
\textbf{Writing – Original Draft Preparation:} Hlynur Davíð Hlynsson
\newline \newline
\textbf{Writing – Review \& Editing:} Hlynur Davíð Hlynsson, Laurenz Wiskott, Tobias Glasmachers, Merlin Schüler, Robin Schiewer

\nolinenumbers


\begin{thebibliography}{10}


\bibitem{lecun2004learning}
LeCun Y, Huang FJ, Bottou L, et~al.
\newblock Learning methods for generic object recognition with invariance to
  pose and lighting.
\newblock In: CVPR (2). Citeseer; 2004. p. 97--104.

\bibitem{schuler2018gradient}
Sch{\"u}ler M, Hlynsson HD, Wiskott L.
\newblock Gradient-based training of slow feature analysis by differentiable
  approximate whitening.
\newblock In: Asian Conference on Machine Learning; 2019. p. 316--331.

\bibitem{hadsell2006dimensionality}
Hadsell R, Chopra S, LeCun Y.
\newblock Dimensionality reduction by learning an invariant mapping.
\newblock In: 2006 IEEE Computer Society Conference on Computer Vision and
  Pattern Recognition (CVPR'06). vol.~2. IEEE; 2006. p. 1735--1742.

\bibitem{wang2018toybox}
Wang X, Ma T, Ainooson J, Cha S, Wang X, Molla A, et~al.
\newblock The Toybox Dataset of Egocentric Visual Object Transformations.
\newblock arXiv preprint arXiv:180606034. 2018;.

\bibitem{corneil2018efficient}
Corneil D, Gerstner W, Brea J.
\newblock Efficient model-based deep reinforcement learning with variational
  state tabulation.
\newblock In: International Conference on Machine Learning; 2018. p. 1049--1058.


\bibitem{hamrick2019analogues}
Hamrick JB.
\newblock Analogues of mental simulation and imagination in deep learning.
\newblock Current Opinion in Behavioral Sciences. 2019;29:8--16.



\bibitem{tamar2016value}
Tamar A, Wu Y, Thomas G, Levine S, Abbeel P.
\newblock Value iteration networks.
\newblock In: Advances in Neural Information Processing Systems; 2016. p. 2154--2162.

\bibitem{srinivas2018universal}
Srinivas A, Jabri A, Abbeel P, Levine S, Finn C.
\newblock Universal planning networks.
\newblock In: International Conference on Machine Learning; 2018. p. 4732--4741.

\bibitem{hafner2018learning}
Hafner D, Lillicrap T, Fischer I, Villegas R, Ha D, Lee H, et~al.
\newblock Learning latent dynamics for planning from pixels.
\newblock In: International Conference on Machine Learning; 2018. p. 2555-2565.

\bibitem{henaff2017model}
Henaff M, Whitney WF, LeCun Y.
\newblock Model-based planning with discrete and continuous actions.
\newblock arXiv preprint arXiv:170507177. 2017;.

\bibitem{chua2018deep}
Chua K, Calandra R, McAllister R, Levine S.
\newblock Deep reinforcement learning in a handful of trials using
  probabilistic dynamics models.
\newblock In: Advances in Neural Information Processing Systems; 2018. p.
  4754--4765.

\bibitem{gal2016improving}
Gal Y, McAllister R, Rasmussen CE.
\newblock Improving PILCO with Bayesian neural network dynamics models.
\newblock In: Data-Efficient Machine Learning workshop, ICML. vol.~4; 2016.

\bibitem{gelada2019deepmdp}
Gelada C, Kumar S, Buckman J, Nachum O, Bellemare MG.
\newblock DeepMDP: Learning Continuous Latent Space Models for Representation
  Learning.
\newblock In: International Conference on Machine Learning; 2019. p. 2170--2179.

\bibitem{oh2015action}
Oh J, Guo X, Lee H, Lewis RL, Singh S.
\newblock Action-conditional video prediction using deep networks in atari
  games.
\newblock In: Advances in neural information processing systems; 2015. p.
  2863--2871.

\bibitem{ha2018world}
Ha D, Schmidhuber J.
\newblock World models.
\newblock arXiv preprint arXiv:180310122. 2018;.

\bibitem{hinton2015distilling}
Hinton G, Vinyals O, Dean J.
\newblock Distilling the knowledge in a neural network.
\newblock arXiv preprint arXiv:150302531. 2015;.

\bibitem{richthofer2015predictable}
Richthofer S, Wiskott L.
\newblock Predictable feature analysis
\newblock In: 2015 IEEE 14th International Conference on Machine Learning
 and Applications (ICMLA). IEEE; 2015. p. 190--196.

\bibitem{bucilua2006model}
Bucilu{ǎ} C, Caruana R, Niculescu-Mizil A.
\newblock Model compression.
\newblock In: Proceedings of the 12th ACM SIGKDD international conference on
  Knowledge discovery and data mining. ACM; 2006. p. 535--541.

\bibitem{vondrick2016anticipating}
Vondrick C, Pirsiavash H, Torralba A.
\newblock Anticipating visual representations from unlabeled video.
\newblock In: Proceedings of the IEEE Conference on Computer Vision and Pattern
  Recognition; 2016. p. 98--106.

\bibitem{kurutach2018learning}
Kurutach T, Tamar A, Yang G, Russell SJ, Abbeel P.
\newblock Learning Plannable Representations With Causal {InfoGAN}.
\newblock In: Advances in Neural Information Processing Systems; 2018. p.
  8733--8744.

\bibitem{goodfellow2014generative}
Goodfellow I, Pouget-Abadie J, Mirza M, Xu B, Warde-Farley D, Ozair S, et~al.
\newblock Generative adversarial nets.
\newblock In: Advances in neural information processing systems; 2014. p.
  2672--2680.

\bibitem{chen2016infogan}
Chen X, Duan Y, Houthooft R, Schulman J, Sutskever I, Abbeel P.
\newblock {InfoGAN}: Interpretable representation learning by information
  maximizing generative adversarial nets.
\newblock In: Advances in neural information processing systems; 2016. p.
  2172--2180.

\bibitem{jeannerod2003action}
Jeannerod M, Arbib M.
\newblock Action monitoring and forward control of movements.
\newblock The Handbook of Brain Theory and Neural Networks,. 2003; p. 83--85.

\bibitem{sharif2014cnn}
Sharif~Razavian A, Azizpour H, Sullivan J, Carlsson S.
\newblock {CNN} features off-the-shelf: an astounding baseline for recognition.
\newblock In: Proceedings of the IEEE conference on computer vision and pattern
  recognition workshops; 2014. p. 806--813.

\bibitem{nilsson2014principles}
Nilsson NJ.
\newblock Principles of artificial intelligence.
\newblock Morgan Kaufmann; 2014.

\bibitem{simonyan2014very}
Simonyan K, Zisserman A.
\newblock Very deep convolutional networks for large-scale image recognition.
\newblock arXiv preprint arXiv:14091556. 2014;.

\bibitem{belkin2003laplacian}
Belkin M, Niyogi P.
\newblock Laplacian eigenmaps for dimensionality reduction and data
  representation.
\newblock Neural computation. 2003;15(6):1373--1396.

\bibitem{russell2016artificial}
Russell SJ, Norvig P.
\newblock Artificial intelligence: a modern approach.
\newblock Malaysia; Pearson Education Limited,; 2016.

\bibitem{sun2014deep}
Sun Y, Chen Y, Wang X, Tang X.
\newblock Deep learning face representation by joint
  identification-verification.
\newblock In: Advances in neural information processing systems; 2014. p.
  1988--1996.

\bibitem{schroff2015facenet}
Schroff F, Kalenichenko D, Philbin J.
\newblock Facenet: A unified embedding for face recognition and clustering.
\newblock In: Proceedings of the IEEE conference on computer vision and pattern
  recognition; 2015. p. 815--823.

\bibitem{harwood2017smart}
Harwood B, Kumar B, Carneiro G, Reid I, Drummond T, et~al.
\newblock Smart mining for deep metric learning.
\newblock In: Proceedings of the IEEE International Conference on Computer
  Vision; 2017. p. 2821--2829.

\bibitem{denton2017unsupervised}
Denton EL, et~al.
\newblock Unsupervised learning of disentangled representations from video.
\newblock In: Advances in neural information processing systems; 2017. p.
  4414--4423.

\bibitem{goroshin2015learning}
Goroshin R, Mathieu MF, LeCun Y.
\newblock Learning to linearize under uncertainty.
\newblock In: Advances in Neural Information Processing Systems; 2015. p.
  1234--1242.

\bibitem{NIPS2019_8413}
Xu D, Mart\'{\i}n-Mart\'{\i}n R, Huang DA, Zhu Y, Savarese S, Fei-Fei LF.
\newblock Regression Planning Networks.
\newblock In: Wallach H, Larochelle H, Beygelzimer A, d\textquotesingle
  Alch\'{e}-Buc F, Fox E, Garnett R, editors. Advances in Neural Information
  Processing Systems 32. Curran Associates, Inc.; 2019. p. 1319--1329.
\newblock Available from:
  \url{http://papers.nips.cc/paper/8413-regression-planning-networks.pdf}.

\bibitem{escalante2013solve}
Escalante-B AN, Wiskott L.
\newblock How to solve classification and regression problems on high-dimensional data with a supervised extension of slow feature analysis.
\newblock The Journal of Machine Learning Research. 2013;14(1):3683--3719.

\bibitem{garcia1989model}
Garcia CE, Prett DM, Morari M.
\newblock Model predictive control: theory and practice—a survey.
\newblock Automatica. 1989;25(3):335--348.

\bibitem{chollet2015keras}
Chollet F, et~al.. Keras; 2015.
\newblock \url{https://keras.io}.

\bibitem{dozat2016incorporating}
Dozat T.
\newblock Incorporating {N}esterov momentum into {ADAM}. 2016;.

\bibitem{deng2009imagenet}
Deng J, Dong W, Socher R, Li LJ, Li K, Fei-Fei L.
\newblock ImageNet: A large-scale hierarchical image database.
\newblock In: 2009 IEEE conference on computer vision and pattern recognition.
  Ieee; 2009. p. 248--255.

\bibitem{barreto2017successor}
Barreto A, Dabney W, Munos R, Hunt JJ, Schaul T, van Hasselt HP, et~al.
\newblock Successor features for transfer in reinforcement learning.
\newblock In: Advances in neural information processing systems; 2017. p.
  4055--4065.
  

\bibitem{xu1029regression}
Xu D, Mart\'{\i}n-Mart\'{\i}n R, Huang DA, Zhu Y, Savarese S, Fei-Fei LF.
\newblock Regression Planning Networks.
\newblock In: Wallach H, Larochelle H, Beygelzimer A, d\textquotesingle
  Alch\'{e}-Buc F, Fox E, Garnett R, editors. Advances in Neural Information
  Processing Systems 32. Curran Associates, Inc.; 2019. p. 1319--1329.
\newblock Available from:
  \url{http://papers.nips.cc/paper/8413-regression-planning-networks.pdf}.
  

\bibitem{wiskott2002slow}
Wiskott L, Sejnowski TJ.
\newblock Slow feature analysis: Unsupervised learning of invariances.
\newblock Neural computation. 2002;14(4):715--770.

\bibitem{sprekeler2011relation}
Sprekeler H.
\newblock On the relation of slow feature analysis and laplacian eigenmaps.
\newblock Neural computation. 2011;23(12):3287--3302.


\bibitem{du2019good}
Du SS, Kakade SM, Wang R, Yang LF.
\newblock Is a Good Representation Sufficient for Sample Efficient
  Reinforcement Learning?
\newblock In: International Conference on Learning Representations; 2019.

\bibitem{masci2011stacked}
Masci, Jonathan and Meier, Ueli and Cire{\c{s}}an, Dan and Schmidhuber, J{\"u}rgen
\newblock Stacked convolutional auto-encoders for hierarchical feature extraction
\newblock In: International conference on artificial neural networks; 2011. p. 52--59.

\bibitem{baselines}
Dhariwal P, Hesse C, Klimov O, Nichol A, Plappert M, Radford A, et~al.. OpenAI
  Baselines; 2017.
\newblock \url{https://github.com/openai/baselines}.


\bibitem{stable-baselines}
Hill, Ashley and Raffin, Antonin and Ernestus, Maximilian and Gleave, Adam and Kanervisto, Anssi and Traore, Rene and Dhariwal, Prafulla and Hesse, Christopher and Klimov, Oleg and Nichol, Alex and Plappert, Matthias and Radford, Alec and Schulman, John and Sidor, Szymon and Wu, Yuhuai. Stable Baselines; 2018.
\newblock \url{https://github.com/hill-a/stable-baselines}

\bibitem{saphal2020seerl}
Saphal R, Ravindran B, Mudigere D, Avancha S, Kaul B.
\newblock SEERL: Sample Efficient Ensemble Reinforcement Learning.
\newblock arXiv preprint arXiv:200105209. 2020;.

\bibitem{wang2016sample}
Wang Z, Bapst V, Heess N, Mnih V, Munos R, Kavukcuoglu K, et~al.
\newblock Sample efficient actor-critic with experience replay.
\newblock In: International Conference on Learning Representations; 2016. 

\bibitem{icpram19}
Hlynsson HD,  ANE, Wiskott L.
\newblock Measuring the Data Efficiency of Deep Learning Methods.
\newblock In: Proceedings of the 8th International Conference on Pattern
  Recognition Applications and Methods - Volume 1: ICPRAM,. INSTICC.
  SciTePress; 2019. p. 691--698.

\bibitem{NIPS2018_8044}
Buckman J, Hafner D, Tucker G, Brevdo E, Lee H.
\newblock Sample-Efficient Reinforcement Learning with Stochastic Ensemble
  Value Expansion.
\newblock In: Bengio S, Wallach H, Larochelle H, Grauman K, Cesa-Bianchi N,
  Garnett R, editors. Advances in Neural Information Processing Systems 31.
  Curran Associates, Inc.; 2018. p. 8224--8234.
\newblock Available from:
  \url{http://papers.nips.cc/paper/8044-sample-efficient-reinforcement-learning-with-stochastic-ensemble-value-expansion.pdf}.
  
 
\bibitem{10.1007/978-3-030-30179-8_15}
Hlynsson HD, Wiskott L.
\newblock Learning Gradient-Based ICA by Neurally Estimating Mutual
  Information.
\newblock In: Benzm{\"u}ller C, Stuckenschmidt H, editors. KI 2019: Advances in
  Artificial Intelligence. Cham: Springer International Publishing; 2019. p.
  182--187.

\bibitem{mnih2013playing}
Mnih V, Kavukcuoglu K, Silver D, Graves A, Antonoglou I, Wierstra D, et~al.
\newblock Playing atari with deep reinforcement learning.
\newblock In: NIPS Deep Learning Workshop; 2013.


\bibitem{schulman2017proximal}
Schulman J, Wolski F, Dhariwal P, Radford A, Klimov O.
\newblock Proximal policy optimization algorithms.
\newblock arXiv preprint arXiv:170706347. 2017;.

\bibitem{kingma2013auto}
Kingma, Diederik P and Welling, Max
\newblock Auto-encoding Variational Bayes
\newblock arXiv preprint arXiv:1312.6114. 2013;.

\bibitem{tieleman2012lecture}
Tieleman T, Hinton G.
\newblock Lecture 6.5-rmsprop: Divide the gradient by a running average of its
  recent magnitude.
\newblock COURSERA: Neural networks for machine learning. 2012;4(2):26--31.


\bibitem{cuccu2018playing}
Cuccu G, Togelius J, Cudré-Mauroux P.
\newblock Playing atari with six neurons.
\newblock In: Proceedings of the 18th International Conference on Autonomous Agents and MultiAgent Systems; 2019. p. 998--1006.


\bibitem{savinov2018semi}
Savinov N, Dosovitskiy A, Koltun V.
\newblock Semi-parametric topological memory for navigation.
\newblock In: International Conference on Learning Representations; 2018.

\bibitem{liu2020hallucinative}
Liu K, Kurutach T, Tung C, Abbeel P, Tamar A.
\newblock Hallucinative Topological Memory for Zero-Shot Visual Planning.
\newblock arXiv preprint arXiv:200212336. 2020;

\bibitem{ebert2018visual}
Ebert F, Finn C, Dasari S, Xie A, Lee A, Levine S.
\newblock Visual foresight: Model-based deep reinforcement learning for
  vision-based robotic control.
\newblock arXiv preprint arXiv:181200568. 2018;.

\bibitem{hamilton2014efficient}
Hamilton W, Fard MM, Pineau J
\newblock Efficient learning and planning with compressed predictive states.
\newblock The Journal of Machine Learning Research. 2014;15(1):3395--3439.

\bibitem{wang2019learning}
Wang, A., Kurutach, T., Liu, K., Abbeel, P., Tamar, A.
\newblock 
Learning robotic manipulation through visual planning and acting
\newblock
arXiv preprint arXiv:1905.04411. 2019;

\bibitem{cassandra1998survey}
Cassandra AR.
\newblock A survey of POMDP applications.
\newblock Working notes of AAAI 1998 fall symposium on planning with partially observable Markov decision processes; 1998;1724.

\bibitem{mcinnes2018umap}
McInnes, Leland and Healy, John and Saul, Nathaniel and Gro{\ss}berger, Lukas
\newblock UMAP: Uniform Manifold Approximation and Projection
\newblock Journal of Open Source Software. 2018; 3(29):861


\end{thebibliography}
\end{document}